\documentclass{article}

\usepackage{subcaption}
\usepackage{times}
\usepackage{url}            %
\usepackage{booktabs}       %
\usepackage{amsfonts}       %
\usepackage{nicefrac}       %
\usepackage{microtype}      %
\usepackage{mkolar_definitions}
\usepackage{xcolor}
\usepackage{xspace}
\usepackage{multirow}
\usepackage{amsmath}
\usepackage{algorithm}
\usepackage{algorithmic}
\usepackage{color}
\usepackage{authblk}
\usepackage{color, colortbl}
\usepackage{enumitem}
\usepackage{comment}
\usepackage{bm}
\usepackage{subcaption}
\usepackage{multirow}
\usepackage{float}

\newcommand{\agl}{\textbf{ELEGANT\,}}

\newcommand{\KL}{\mathrm{KL}}

\newcommand{\data}{\mathrm{data}}

\newcommand{\tar}{\mathrm{tar}}
\newcommand{\reef}{\mathrm{ref}}
\newcommand{\fun}{\mathrm{data}}
\newcommand{\fix}{\mathrm{fix}}
\newcommand{\ini}{\mathrm{ini}}
\newcommand{\pre}{\mathrm{data}}

\newcommand{\condition}{\mathrm{con}}

\definecolor{Gray}{gray}{0.9}

\usepackage{xcolor}
\usepackage{tabularray}

\usepackage{microtype}
\usepackage{graphicx}
\usepackage{booktabs} %

\usepackage{hyperref}

 \usepackage[accepted]{icml2023}

\usepackage{amsmath}
\usepackage{amssymb}
\usepackage{mathtools}
\usepackage{amsthm}

\usepackage[capitalize,noabbrev]{cleveref}

\usepackage[textsize=tiny]{todonotes}

\icmltitlerunning{Fine-Tuning of Continuous-Time Diffusion Models as Entropy-Regularized Control}

\begin{document}

\twocolumn[
\icmltitle{Fine-Tuning of Continuous-Time Diffusion Models \\ as Entropy-Regularized Control}

\icmlsetsymbol{equal}{*}
\begin{icmlauthorlist}

\icmlauthor{Masatoshi Uehara}{equal,gen}
\icmlauthor{Yulai Zhao}{equal,pre}
\icmlauthor{Kevin Black}{ber}
\icmlauthor{Ehsan Hajiramezanali}{gen}
\icmlauthor{Gabriele Scalia}{gen}
\icmlauthor{Nathaniel Lee Diamant}{gen}
\icmlauthor{Alex M Tseng}{gen}
\icmlauthor{Tommaso Biancalani$^{\dagger}$}{gen}
\icmlauthor{Sergey Levine$^{\dagger}$}{ber}
\end{icmlauthorlist}

\icmlaffiliation{gen}{Genentech}
\icmlaffiliation{pre}{Princeton University }
\icmlaffiliation{ber}{University of California, Berkeley}

\icmlcorrespondingauthor{Tommaso Biancalani }{biancalt@gene.com}

\icmlcorrespondingauthor{Sergey Levine}{sergey.levine@berkeley.edu}

\icmlkeywords{Machine Learning, ICML}

\vskip 0.3in
]

\printAffiliationsAndNotice{\icmlEqualContribution} %

\begin{abstract}
{ Diffusion models excel at capturing complex data distributions, such as those of natural images and proteins. While diffusion models are trained to represent the distribution in the training dataset, we often are more concerned with other properties, such as the aesthetic quality of the generated images or the functional properties of generated proteins. Diffusion models can be finetuned in a goal-directed way by maximizing the value of some reward function (e.g., the aesthetic quality of an image). However, these approaches may lead to reduced sample diversity, significant deviations from the training data distribution, and even poor sample quality due to the exploitation of an imperfect reward function. The last issue often occurs when the reward function is a learned model meant to approximate a ground-truth ``genuine'' reward, as is the case in many practical applications.
These challenges, collectively termed ``reward collapse,'' pose a substantial obstacle. To address this reward collapse,
we frame the finetuning problem as entropy-regularized control against the pretrained diffusion model, i.e., directly optimizing entropy-enhanced rewards with neural SDEs. We present theoretical and empirical evidence that demonstrates our framework is capable of efficiently generating diverse samples with high genuine rewards, mitigating the overoptimization of imperfect reward models.} 
\end{abstract}

\vspace{-7mm}
\section{Introduction}
 \vspace{-2mm}

\label{sec:intro}

Diffusion models have gained widespread adoption as effective tools for modeling complex distributions~\citep{sohl2015deep,song2020score,ho2020denoising}. These models have demonstrated state-of-the-art performance in various domains such as image generation and biological sequence generation~\citep{jing2022torsional,wu2022diffusion}. While diffusion models effectively capture complex data distributions, our primary goal frequently involves acquiring a finely tuned sampler customized for a specific task using the pre-trained diffusion model as a foundation. For instance, in image generation, we might like to fine-tune diffusion models to enhance aesthetic quality. In biology, we might aim to improve bioactivity. Recent endeavors have pursued this objective through reinforcement learning (RL) \citep{fan2023dpok,black2023training} as well as direct backpropagation through differentiable reward functions \citep{clark2023directly,prabhudesai2023aligning}. Such reward functions are typically learned models meant to approximate a ground-truth ``genuine'' reward; e.g., an aesthetic classifier is meant to approximate the true aesthetic preferences of human raters.

\begin{figure}[!t]
    \centering
    \includegraphics[width = 1.0\linewidth]{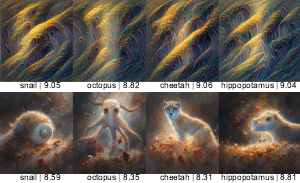}
    \caption{\textbf{Mitigating nominal reward exploitation with entropy-regularized control.} Diffusion models fine-tuned in a goal-directed manner can produce images (top) with high nominal reward values such as aesthetic scores. However, these images lack realism because the naive fine-tuning process is not incentivized to stay close to the pre-trained data distribution. Our approach (bottom) mitigates this issue via entropy-regularized stochastic optimal control. } 
    \label{fig:reward}
\end{figure}

While these methods allow us to generate samples with high ``nominal" (approximate) rewards, they often suffer from reward collapse. Reward collapse manifests as fine-tuned models produce the same samples with low genuine rewards, as illustrated in Figure \ref{fig:reward}. This issue arises because nominal rewards are usually learned from a finite number of training data to approximate the genuine reward function, meaning that they are accurate only within their training data distribution. Consequently, fine-tuning methods quickly exploit nominal rewards by moving beyond the support of this distribution.
Even ignoring nominal reward exploitation, reward collapse is problematic because diversity is, in itself, a desirable property of generative models. For example, we typically expect generative models to generate diverse candidates of protein sequences to run good wet lab experiments \citep{watson2022learning}.

Our goal is to develop a principled algorithmic framework for fine-tuning diffusion models that both optimizes a user-specified reward function and captures a diverse distribution that stays close to the training data, thus alleviating reward collapse issues. To achieve this, we frame the fine-tuning of diffusion models as an entropy-regularized control problem. It is known that diffusion models can be formulated as stochastic differential equations (SDEs) with a drift term and a diffusion term \citep{song2020score}. Based on this formulation, in a fine-tuning step, we consider solving stochastic control by neural SDEs in a computationally efficient manner. Here, we introduce a loss that combines a terminal reward with entropy regularization against the pre-trained diffusion model, with respect to both a drift term and an initial distribution. This entropy-regularization term enables us to maintain the bridges (i.e., the posterior distributions of trajectories conditioned on a terminal point) of pre-trained diffusion models, akin to bridge-matching generative models \citep{shi2023diffusion}, such that the fine-tuned diffusion model avoids deviating too much from the pre-trained diffusion model.  

Notably, we formally show that the fine-tuned SDE, optimized for both the drift term and initial distribution, can produce specific distributions with high diversity and high nominal rewards that are within the support of their training data distribution. Hence, our approach effectively mitigates the reward-collapse problem since nominal rewards accurately approximate genuine rewards in that region.

Our contribution can be summarized as follows: we introduce a computationally efficient, theoretically and empirically supported method for fine-tuning diffusion models: \agl (finE-tuning doubLe Entropy reGulArized coNTrol) that excels at generating samples with both high genuine rewards and a high degree of diversity. While existing techniques in image generation~\citep{fan2023dpok,prabhudesai2023aligning,clark2023directly} include components for mitigating reward collapse, we demonstrate stronger theoretical support (among methods that directly backpropagate through differentiable rewards) and superior empirical performance (compared to a KL-penalized RL approach). Additionally, unlike prior work, we apply our method to both image generation and biological sequence generation, demonstrating its effectiveness across multiple domains.

\vspace{-3mm}
\section{Related Works}
\vspace{-2mm}

We provide an overview of related works. We leave the discussion of our work, including fine-tuning LLMs, sampling with control methods, and MCMC methods to Appendix~\ref{sec:related}. 

\vspace{-2mm}
\paragraph{Diffusion models.}  Denoising diffusion probabilistic models (DDPMs)
create a dynamic stochastic transport using SDEs, where the drift aligns with a specific score function \citep{song2020denoising,ho2020denoising}. The impressive performance of DDPMs has spurred the recent advancements in bridge (flow)-matching techniques, which construct stochastic transport through SDEs with drift terms aligned to specific bridge functions \citep{liu2022flow,shi2023diffusion,tong2023conditional,lipman2022flow,somnath2023aligned,liu20232,delbracio2023inversion}.    
For a more comprehensive review, we direct readers to \citet{shi2023diffusion}. Note our that fine-tuning methods are applicable to any model with diffusion terms, but are not suitable for purely ODE-based models  \citep{papamakarios2021normalizing}.

\vspace{-2mm}
\paragraph{Guidance.}
\citet{dhariwal2021diffusion} introduced classifier-based guidance, an inference-time technique for steering diffusion samples towards a particular class. More generally, guidance uses an auxiliary differentiable objective (e.g., a neural network) to steer diffusion samples towards a desired property \citep{graikos2022diffusion,bansal2023universal}. In our experiments, we show that our fine-tuning technique outperforms a guidance baseline that uses the gradients of the reward model to steer the pre-trained diffusion model toward high-reward regions.

\vspace{-2mm}
\paragraph{Fine-tuning as RL/control.}
\citet{lee2023aligning,wu2023better} employ supervised learning techniques to optimize reward functions, while \citet{black2023training,fan2023dpok} employ an RL-based method to achieve a similar goal. \citet{clark2023directly,xu2023imagereward,prabhudesai2023aligning} present a fine-tuning method that involve direct backpropagation with respect to rewards, which bears some resemblance to our work. Nevertheless, there are several notable distinctions between our approaches. Specifically, we incorporate an entropy-regularization term and also learn an initial distribution, both of which play a critical role in targeting the desired distribution. We present novel theoretical results that demonstrate the benefits of our approach, and we provide empirical evidence that our method more efficiently mitigates reward collapse.

\vspace{-2mm}
\section{Preliminaries}
\vspace{-2mm}

We briefly review current continuous-time diffusion models.  
A diffusion model is described by the following SDE:  
\begin{align}\label{eq:fun}
     d \xb_t = f(t,\xb_t) d t +\sigma(t) d \wb_t, \quad \xb_0\sim \nu_{\ini} \in \Delta(\RR^d),
\end{align}
where $f:[0,T] \times \RR^d \to \RR^d$ is a drift coefficient, and $\sigma:[0,T]\to \RR_{>0}$ is a diffusion coefficient associated with a $d$-dimensional Brownian motion $\wb_t$, and $\nu_{\ini}$ is an initial distribution such as a Gaussian distribution. Note that many papers use the opposite convention, with $t=T$ corresponding to the initial distribution and $t=0$ corresponding to the data.
When training diffusion models, the goal is to learn $f(t,x_t)$ from the data at hand so that the generated distribution from the SDE \eqref{eq:fun} corresponds to the data distribution through score matching \citep{song2020denoising} or bridge/flow matching \citep{liu2022let}. For details, refer to Appendix~\ref{subsec:continuous}. 

In our work, we focus on scenarios where we have such a pre-trained diffusion model (i.e., a pre-trained SDE). Denoting the density at time $T$ induced by the pre-trained SDE in \pref{eq:fun} as $p_{\fun} \in \Delta(\RR^d)$, this $p_{\fun}$ captures the intricate structure of the data distribution. In image generation, $p_{\fun}$ captures the structure of natural images, while in biological sequence generation, it captures the biological space. 

\vspace{-2mm}
\paragraph{Notation.}
\vspace{-2mm}

We often consider a measure $\PP$ induced by an SDE on $\Ccal:=C([0,T],\RR^d)$ where $C([0,T],\RR^d)$ is the whole set of continuous functions mapping from $[0,T]$ to $\RR^d$ \citep{karatzas2012brownian}. The notation $\EE_{\PP}[f(x_{0:T})]$ means that the expectation is taken for $f(\cdot)$ w.r.t. $\PP$. We denote $\PP_t$ as the marginal distribution over $\RR^d$ at time $t$, $\PP_{s,t}(x_s,x_t)$ the joint distribution over $\RR^d$ time $s$ and $t$, and $\PP_{s|t}(x_s|x_t)$ the conditional distribution at time $s$ given time $t$. We also denote the distribution of the process pinned down at an initial and terminal point $x_0,x_T$ by $\PP_{\cdot|0,T}(\cdot |x_0, x_T)$ (we similarly define $\PP_{\cdot|T}(\cdot |x_T)$). With a slight abuse of notation, we exchangeably use distributions and densities \footnote{We sometimes denote densities such as $d \PP_{T}/d\mu$ by just $\PP_{T}$ where $\mu$ is Lebesgue measure. }

We defer all proofs to Appendix~\ref{sec:proof}.

\vspace{-2mm}
\section{Desired Properties for Fine-Tuning}\label{sec:desired}
\vspace{-2mm}

In this section, we elucidate the desired properties for methods that fine-tune diffusion models.

In the presence of a reward function $r:\RR^d \to \RR$, such as aesthetic quality in image generation or bioactivity in biological sequence generation, our aim might be to fine-tune a pre-trained diffusion model so as to maximize this reward function, for example to generate images that are more aesthetically pleasing.

{ However, the ``genuine'' reward function (e.g., a true human rating of aesthetic appearance) is usually unknown, and instead a computational proxy must be learned from data --- typically from the same or a similar distribution as the pre-training data for the diffusion model. As a result, while $r(x)$ may be close to the genuine reward function within the support of $p_{\data}$, it might not be accurate outside of this domain. More formally, by denoting the genuine reward by $r^{\star}$, a nominal reward $r$  is typically learned as   
\begin{align*}\textstyle 
    r= \argmin_{r' \in \Fcal}\sum_{i} \{r^{\star}(x^{(i)})-r'(x^{(i)}) \}^2] ,
\end{align*} 
where $\{x^{(i)},r^{\star}(x^{(i)})\}_{i=1}^n$ is a dataset, and $\Fcal$ is a function class (e.g., neural networks) mapping from $\RR^d$ to $\RR$. Under mild conditions, it has been shown that in high probability, the mean square error on $p_{\fun}$ is small, i.e.,  
\begin{align*}
    \EE_{x \sim p_{\fun}}[\{r^{\star}(x) - r(x) \}^2]=O(\sqrt{\mathrm{Cap}(\Fcal)/n}),
\end{align*}
where $\mathrm{Cap}(\Fcal)$ is a capacity of $\Fcal$ \citep{wainwright2019high}. However, this does not hold outside of the support of $p_{\fun}$. 
} 

Taking this into account, we aim to fine-tune a diffusion model in a way that preserves three essential properties: (a1) the ability to generate samples with high rewards, (a2) maintaining the diversity of samples,
and (a3) ensuring sufficient proximity to the initial pre-trained diffusion \eqref{eq:fun}.
In particular, (a3) helps avoid overoptimization because learned reward functions tend to be accurate on the support of $p_{\data}$.

To accomplish this, we consider the optimization problem: 
\begin{align}\label{eq:final}
   p_{\tar}=\argmax_{p \in \Delta(\RR^d) }\underbrace{\EE_{x\sim p}[{r(x)}]}_{\Psi(1)} -\alpha {\underbrace{\KL(p\|p_{\fun})}_{\Psi(2)} },
\end{align}
where $\alpha \in \RR_{>0}$ is a hyperparameter. The initial reward term $\Psi(1)$ is intended to uphold the property (a1), while the second entropy term $\Psi(2)$  is aimed at preserving the property (a2). To achieve (a2), one might consider just adding a relative entropy against a uniform distribution. However, in our case, we add the relative entropy against $p_{\fun}$ to maintain property (a3).

It can be shown that the target distribution in \eqref{eq:final} takes the following analytical form: 
\begin{align}\label{eq:target}
     p_{\tar}(x)= \exp (r(x)/\alpha) )p_{\fun}(x)/C_{\tar} ,
\end{align}
where $C_{\tar}$ is a normalizing constant. Therefore, the aim of our method is to provide a tractable and theoretically principled way to emulate $ p_{\tar}$ as a fine-tuning step. 

\vspace{-2mm}
\subsection{Importance of KL Regularization}\label{subsec:kl_term}
\vspace{-2mm}

Before explaining our approach to sample from $ p_{\tar}$, we elucidate the necessity of incorporating entropy regularization term in \eqref{eq:final}. This can be seen by examining the limit cases as $\alpha$ tends towards $0$ and when we fix $\alpha=0$ a priori. To be more precise, as $\alpha$ approaches zero, $p_{\tar}$ tends to converge to a Dirac delta distribution at $x^{\star}_{\tar}$, defined by:
\begin{align*}\textstyle 
   x^{\star}_{\tar}=\argmax_{x \in \RR^d:p_{\fun}(x)>0}r(x). 
\end{align*}
This $x^{\star}_{\tar}$ represents an optimal $x$ within the support of $p_{\fun}$. Conversely, if we directly solve \eqref{eq:final}  with $\alpha=0$, we may venture beyond the support:
\begin{align*}\textstyle 
   x^{\star}=\argmax_{x \in \RR^d}r(x). 
\end{align*}
This implies that the generated samples might no longer adhere to the characteristics of natural images in image generation or biological sequences within the biological space. As we mentioned, since $r(x)$ is typically a learned reward function from the data, it won't be accurate outside of the support of $p_{\fun}(x)$. Hence, $x^{\star}$ would not have a high genuine reward, which results in ``overoptimization''. For example, this approach results in the unnatural but high nominal reward images in \pref{fig:reward}.

\begin{remark}[Failure of na\"ive approach]\label{rem:naive}
The score function of $p_{\tar}(\cdot)$ w.r.t. $x$ is given by $\nabla r(x)/\alpha + \nabla \log p_{\pre}(x)$. Therefore, some readers might question whether we can straightforwardly use:
\begin{align*}
     d \xb_t = \{f(t,\xb_t)+ \nabla r(\xb_t)/\alpha\} d t +\sigma(t) d \wb_t
\end{align*}
However, there is no validity of this approach because we cannot simply use $\nabla r(x)$ for time $t\neq T$. We will see a correct approach in Lemma~\ref{lem:optimal_drift}.
\end{remark}

\vspace{-2mm}
\section{Entropy-Regularized Control with Pre-Trained Models}
\vspace{-2mm}

We show how to sample from the target distribution $p_{\tar}$ using entropy-regularized control and a pre-trained model.

\vspace{-2mm}
\subsection{Stochastic Control Formulation}
\vspace{-2mm}

To fine-tune diffusion models, we consider the following SDE by adding an additional drift term $u$ and changing the initial distribution of \eqref{eq:fun}:
\begin{align}\label{eq:new_SDE}
    d \xb_t = \{f(t,\xb_t) + u(t,\xb_t)\}d t + \sigma(t)d \wb_t, x_0\sim \nu, 
\end{align}
where $u(\cdot,\cdot):[0,T]\times \RR^d \to \RR$ is a drift coefficient we want to learn and $\nu \in \Delta(\RR^d)$ is an initial distribution we want to learn. When $u=0$ and $\nu=\nu_{\ini}$,
this reduces to a pre-trained SDE in \eqref{eq:fun}. Our objective is to select $u$ and $\nu$ in such a way that the density at time $T$, induced by this SDE, corresponds to $p_{\tar}$. 

Now, let's turn our attention to the objective function designed to achieve this objective. Being motivated by \eqref{eq:final}, the objective function we consider is as follows:
\begin{align}\label{eq:obje}
 & u^{\star}, \nu^{\star}= \argmax_{u,\nu} \EE_{\PP^{u,\nu} } [\underbrace{r(\xb_T)}_{(b1)}]\\
 -&\frac{\alpha}{2} \underbrace{\EE_{\PP^{u,\nu} }\left [  \int_{t=0}^T \frac{\|u(t,\xb_t)\|^2}{\sigma^2(t)}dt + \log \left (\frac{\nu(\xb_0)}{\nu_{\ini}(x_0)}\right)\right]}_{(b2) }  \nonumber
\end{align} 
where $\PP^{u,\nu}$ is a measure over $\Ccal$ induced by the SDE \eqref{eq:new_SDE} associated with $(u,\nu)$. Within this equation, component (b1) is introduced to obtain samples with high rewards. This is equal to $\Psi(1)$ in \eqref{eq:final} when $p(\cdot)$ in \eqref{eq:final} comes from  $\PP^{u,\nu}_T$. The component (b2) corresponds to the KL divergence over trajectories:
$\KL(\PP^{u,\nu}(\cdot) \|\PP^{\fun}(\cdot))$ where $\PP^{\fun}$ is a measure over $\Ccal$ induced by the pre-trained SDE \eqref{eq:fun}. This is actually equal to $\Psi(2)$ in \eqref{eq:final} as we will see in the proof of our key theorem soon. 
 
We can derive an explicit expression for the marginal distribution at time $t$ under the distribution over $\Ccal$ induced by the SDE associated with the optimal drift and initial distribution denoted by $\PP^{\star}$ (i.e., $\PP^{u^{\star},\nu^{\star}}$). Here, we define 
the optimal (entropy-regularized) value function as 
\begin{align*}
    v^{\star}_{t}(x) &= \EE_{\PP^{\star}} \left [r(\xb_T)-\frac{\alpha}{2} \int_{k=t}^T \frac{\|u(k,\xb_k)\|^2}{\sigma^2(k)}dk   |\xb_t=x \right]. 
\end{align*}

\begin{theorem}[Induced marginal distribution]\label{thm:important}
 The marginal density at step $t \in [0,T]$ under the diffusion model with a drift term $u^{\star}$ and a diffusion coefficient $\nu^{\star}$ (i.e., $\PP^{\star}_t$) is 
\begin{align}\label{eq:optimal}
\PP^{\star}_t(\cdot)= \exp(v^{\star}_t(x)/\alpha)\PP^{\fun}_t(x)/C_{\tar} 
\end{align}
where $\PP^{\fun}_t(\cdot)$ is a marginal distribution at $t$ of $\PP^{\fun}$. 
\end{theorem}
This marginal density comprises two components: the optimal value function term and the density at time $t$ induced by the pre-trained diffusion model. Note that the normalizing constant $C_{\tar}$ is independent of $t$. 

Crucially, as a corollary, we observe that by generating a sample following the SDE \eqref{eq:new_SDE} with $(u^{\star},\nu^{\star})$, we can sample from the target $p_{\tar}$ at the final time step $T$. Furthermore, we can also determine the explicit form of $\nu^{\star}$.

\begin{corollary}[Justification of control problem]
$$ \textstyle \PP_T(\cdot)=p_{\tar}(\cdot).$$ 
\end{corollary}

\begin{corollary}[Optimal initial distribution]
\begin{align*}
 \textstyle  \nu^{\star}(\cdot)=  \exp(v^{\star}_0(\cdot)/\alpha)\nu_{\ini}(\cdot)/C_{\tar}. 
\end{align*}
\end{corollary}

In the following section, to gain deeper insights, we explore two interpretations.

\vspace{-2mm}
\subsection{Feynman–Kac Formulation}
\vspace{-2mm}

We see an interpretable formulation of the optimal value function. We use this form to learn the optimal initial distribution later in our algorithm and the proof of Theorem~\ref{thm:important}. 

\begin{lemma}[Feynman–Kac Formulation]\label{lem:Feynman}
  \begin{align*}
    \exp\left(\frac{v^{\star}_{t}(x)}{\alpha} \right)= \EE_{\PP^{\pre}}\left [\exp \left (\frac{r(\xb_T)}{\alpha } \right)|\xb_t=x \right].
\end{align*}  
\end{lemma}

This lemma illustrates that the value function at $(t,x)$ is higher when it allows us to hit regions with high rewards at $t=T$ by following the pre-trained diffusion model afterward. Using this optimal value function, we can write the optimal drift $u^{\star}(t,x)$ as $\sigma^2(t)\nabla_x v^{\star}_t(x)/\alpha.$ By plugging Lemma~\ref{lem:Feynman} into $\sigma^2(t) \nabla_x v^{\star}_t(x)/\alpha$, we obtain the following. 

\begin{lemma}[Optimal drift]\label{lem:optimal_drift}
    \begin{align*}
     u^{\star}(t,x)=\sigma^2(t) \nabla_{x} \left\{ \log \EE_{\PP^{\pre}}\left [\exp\left (\frac{r(\xb_T)}{\alpha} \right)|\xb_t=x \right]\right\} . 
    \end{align*}
\end{lemma}
It says the optimal control aims to move the current state $x$ at time $t$ toward a point where it becomes easier to achieve higher rewards after following the pre-trained diffusion. 

Note that this equals $\nabla_x r(x)/\alpha $ when $t=T$, but not at $t<T$. Therefore, the straightforward approach in Remark~\ref{rem:naive} does not work.  With certain approximation, this becomes  
\begin{align*}\textstyle 
        u^{\star}(t,x)\approx \sigma^2(t)/\alpha \EE_{\PP^{\pre}}\left [r(\xb_T)|\xb_t=x \right]. 
\end{align*}
This formulation has a close connection with classifier guidance \citep{dhariwal2021diffusion}. For more details, refer to Appendix~\ref{ape:classfier}.

\vspace{-3mm}
\subsection{Bridge Preserving Property} 
\vspace{-2mm}
We start by exploring more explicit representations of joint and conditional distributions to deepen the understanding of our control problem.

\begin{lemma}[Joint distributions]\label{lem:joint}
Let $0\leq s<t\leq T$. Then, 
    \begin{align}
   \PP^{\star}_{s,t}(x, y) &=\PP^{\pre}_{s,t}(x,y)\exp(v^{\star}_t(y)/\alpha)/C_{\tar},  \nonumber \\ 
   \PP^{\star}_{s|t}(x|y) &= \PP^{\pre}_{s|t}(x|y). \label{eq:posterior}
\end{align}
\end{lemma}

Interestingly, in \pref{eq:posterior}, the posterior distributions of pre-trained SDE and optimal SDE are identical. This property is a result of the entropy-regularized term. This theorem can be generalized further. 

\begin{lemma}[Bridge perseverance]\label{lem:brdige}
Let $\PP^{\star}_{\cdot|T}(\cdot|x_T),\PP^{\pre}_{\cdot|T}(\cdot|x_T)$ be distributions of $\PP^{\star},\PP^{\pre}$ conditioned on states at terminal $T$, respectively. Then, 
    \begin{align*}
          \PP^{\star}_{\cdot|T}(\cdot|x_T)= \PP^{\pre}_{\cdot|T}(\cdot|x_T).
    \end{align*}
\end{lemma}

As an immediate corollary, we also obtain $\PP^{\star}_{\cdot|0,T}(\cdot)= \PP^{\pre}_{\cdot|0,T}(\cdot)$. These posterior distributions are often referred to as bridges. Note that in bridge matching methods, generative models are trained to align the bridge with the reference Brownian bridge while maintaining the initial distribution as $\nu_{\ini}$ and the terminal distribution as $p_{\pre}$ \citep{shi2023diffusion}. Our fine-tuning method can be viewed as a bridge-matching \emph{fine-tuning} approach between $0$ and $T$ while keeping the terminal distribution as $\exp(r(x)/\alpha)p_{\pre}(x)/C_{\tar}$. This bridge-matching property is valuable in preventing samples from going beyond the support of $p_{\pre}$.

\vspace{-3mm}
\section{Learning an Optimal Initial Distribution via Entropy-Regularized Control}

Up to this point, we have illustrated that addressing the stochastic control problem enables the creation of generative models for the target density $p_{\tar}$. Existing literature on neural ODE/SDEs \citep{chen2018neural,li2020scalable,tzen2019theoretical} has established that these control problems can be effectively solved by relying on the expressive power of a neural network, and employing sufficiently small discretization steps. 

Although it seems plausible to employ any neural SDE solver for solving stochastic control problems \eqref{eq:obje}, in typical algorithms, the initial point is fixed. Even when the initial point is unknown, it is commonly assumed to follow a Dirac delta distribution. In contrast, our control problem in Eq.~\eqref{eq:obje} necessitates the learning of a stochastic initial distribution, which can function as a sampler.

A straightforward way involves assuming a Gaussian model with a mean parameterized by a neural network. While this approach is appealingly simple, it may lead to significant misspecification when $\nu^{\star}$ is a multi-modal distribution.

To address this challenge, we once again turn to approximating $\nu^{\star}$ using an SDE, as SDE-induced distributions have the capability to represent intricate multi-modal distributions. We start with a reference SDE over the interval $[-T,0]$:
\begin{align*}
 t \in [-T,0];   d \xb_t = \tilde \sigma(t)d\wb_t,\, \xb_{-T}=x_{\fix},
\end{align*}
such that the distribution at time $0$ follows $\nu_{\ini}$. Given that $\nu_{\ini}$ is typically simple (e.g., $\mathcal{N}(0,\II_d)$), it is usually straightforward to construct such an SDE with a diffusion coefficient  $\tilde \sigma:[0,T]\to \RR$. 

Building upon this baseline SDE, we introduce another SDE over the same interval $[-T,0]$:
\begin{align}\label{eq:new_learn}
    d \xb_t = q(t,\xb_t)\xb_t + \tilde \sigma(t)d\wb_t,\, \xb_{-T}=x_{\fix}. 
\end{align}
where $q:[-T,0]\times \RR^d \to \RR$. This time, we aim to guide a drift coefficient $q$ over this interval $[-T,0]$ such that the distribution at $0$ follows $\nu^{\star}$. Specifically, we formulate the following stochastic control problem:
\begin{align}\label{eq:another_control} \textstyle 
q^{\star}=\argmax_{q}\EE_{\PP^{q}}\left[v^{\star}_0(\xb_0)-\frac{\alpha}{2}\int_{-T}^0 \frac{\|q(t,\xb_t)\|^2}{\tilde \sigma^2(t) }dt  \right],
\end{align}
where $\PP^{q}$ represents the measure induced by the SDE \eqref{eq:new_learn} with a drift coefficient $q$.

\begin{theorem}[Justification of the second control problem]\label{thm:justification}
The marginal density at time $0$ induced by the SDE \eqref{eq:new_learn} with the drift $q^{\star}$, i.e., $\PP^{q^{\star}}_0(\cdot)$, is $\nu^{\star}(\cdot)$
\end{theorem}

This shows that by solving \eqref{eq:another_control} and following the learned SDE from $-T$ to $0$, we can sample from $\nu^{\star}$ at time $0$.

\vspace{-3mm}
\section{Algorithm}
\vspace{-2mm}

\begin{algorithm}[!t]
\caption{\agl (finE-tuning doubLe Entropy reGulArized coNTrol)  }\label{alg:main}
\begin{algorithmic}[1]
  \STATE {\bf Require}: Parameter $\alpha \in \RR^{+}$, a pre-trained diffusion model with drift coefficient $f:[0,T]\times \RR^d\to \RR$ and diffusion coefficient $\sigma:[0,T] \to \RR$, a base coefficient $\tilde \sigma:[-T,0] \to \RR$ and a base initial point $x_{\fix}$.
  
  \STATE Learn an optimal value function at $t=0$ (i.e., $v^{\star}_0$) and denote it by $\hat a:\RR^d \to \RR$ invoking \textbf{Algorithm}~\ref{alg:awr}. 
  \STATE Using a neural SDE solver (\textbf{Algorithm}~\ref{alg:neural}), solve 
  \begin{align}\label{eq:first_stage}\textstyle 
    \hat q=\argmax_{q}\EE_{\PP^{q}}\left[\hat a(\xb_0)-\frac{\alpha}{2} \int_{-T}^0 \frac{\|q(t,\xb_t)\|^2}{\tilde \sigma^2(t) }dt  \right].
\end{align} 
    \STATE Let $\hat \nu$ be a distribution at $t=0$ following the SDE: 
    \begin{align*}
    d \xb_t = \hat q(t,\xb_t)dt + \tilde \sigma(t)d\wb_t,\, x_{-T}=x_{\fix}.
\end{align*}
 \STATE Using a neural SDE solver (\textbf{Algorithm}~\ref{alg:neural}), solve 
 \begin{align}\label{eq:second_stage} \textstyle 
 \hat u = \argmax_{u}\EE_{\PP^{u,\hat \nu} } \left [{r(\xb_T)}- \frac{\alpha}{2}  {\int_{t=0}^T \frac{\|u(t,\xb_t)\|^2}{\sigma^2(t)}dt } \right].
\end{align} 
  \STATE {\bf Output}: Drift coefficients $\hat q,\hat u$
\end{algorithmic}
\end{algorithm}

\begin{algorithm}[!t]
\caption{Fine-Tuned Sampler}\label{alg:sampling}
\begin{algorithmic}[1]
  \STATE  From $-T$ to $0$, follow the SDE: 
  \begin{align*}\textstyle 
    d \xb_t =  \hat q(t,\xb_t)dt + \tilde \sigma(t)d\wb_t,\, \xb_{-T}=x_{\fix}
\end{align*}
\STATE  From $0$ to $T$, follow the SDE: 
  \begin{align*} \textstyle 
    d \xb_t = \{f(t,\xb_t) + \hat u(t,\xb_t)\}dt +  \sigma(t)d\wb_t. 
\end{align*}
 \STATE {\bf Output}: $x_T$
\end{algorithmic}
\end{algorithm}

We are ready to present our method, \agl, which is fully described in Algorithm~\ref{alg:main}. The algorithm consists of 3 steps:
\setlist{nolistsep}
\begin{enumerate}
\item Learn the value function $v^{\star}_0(x)$, which we will discuss in Section~\ref{subsec:value}. 
    \item Solve the stochastic control \eqref{eq:first_stage} with a neural SDE solver using the learned $v^{\star}_0(x)$ in the first step.  
    \item Solve the stochastic control \eqref{eq:second_stage} with a neural SDE
    using the learned $\nu^{\star}$ in the second step (i.e., $\hat \nu$). Compared to \eqref{eq:obje}, we fix the initial distribution as $\hat \nu$. 
\end{enumerate}

For our neural SDE solver, we use a standard oracle in Algorithm~\ref{alg:neural} as in \citet{kidger2021efficient,chen2018neural}. A detailed implementation is described in Appendix~\ref{subsec:neuralSDE}. To solve \eqref{eq:first_stage}, we use the following parametrization: 
\begin{align*}
    & \zb_t:=[\xb^{\top}_t,y_t]^{\top} \in \RR^{d+1} , L:=  -y_0 + \hat a(\xb_0)  , \zb_{\ini}:=x_{\fix}  \\
    &  \bar  f:=\left [q^{\top}, -0.5\alpha\|q\|^2/\tilde \sigma^2   \right]^{\top},\, \bar g := [\tilde \sigma \mathbf{1}_d, 0]^{\top}
\end{align*}
Similarly, to solve \eqref{eq:second_stage}, we can use this solver with the following parametrization: 
\begin{align*}
    & \zb_t:=[\xb^{\top}_t,y_t]^{\top}\in \RR^{d+1}  , L:= -y_T + r(\xb_T) , \zb_{\ini}:=\hat \nu  \\
    &  \bar  f:=\left [\{f + u\}^{\top}, -0.5\alpha \|u\|^2/\sigma^2 \right]^{\top},\,\bar g := [\sigma \mathbf{1}_d, 0]^{\top}. 
\end{align*}
Finally, after learning the drift terms $\hat q$ and $\hat u$, during the sampling phase, we follow the learned SDE (Algorithm~\ref{alg:sampling}).

\begin{algorithm}[!t]
\caption{NeuralSDE Solver }\label{alg:neural}
\begin{algorithmic}[1]
  \STATE {\bf Input}: A diffusion coefficient $\bar g:[0,T] \to \Zcal$, a loss function $L:\Zcal \to \RR$, an initial distribution $\bar \nu$
  \STATE Solve the following and denote the solution by $f^{\dagger} $: 
  \begin{align*}
\argmax_{\bar f:[0,T]\times \Zcal \to \Zcal  }L( \zb_T),
d\zb_t = \bar f(t,\zb_t) dt + \bar g(t) d\wb_t, z_0\sim \zb_{\ini}. 
\end{align*}
  \STATE {\bf Output:} $f^{\dagger}$
\end{algorithmic}
\end{algorithm}

\label{sec:awr}

\vspace{-2mm}
\subsection{Value Function Estimation} \label{subsec:value}
\vspace{-2mm}

In the initial stage of \agl, our objective is to learn $v^{\star}_0(x_0)$. To achieve this, we use the following:
\begin{align*}
v^{\star}_0(x) = \alpha \log( \EE_{\PP^{\pre}}[\exp(r(\xb_T)/\alpha)|\xb_0=x]), 
\end{align*} 
which is obtained as a corollary of Lemma~\ref{lem:Feynman} at $t=0$. Then, by taking a differentiable function class $\Acal:\RR^d \to \RR$, we can use an empirical risk minimization (ERM) algorithm to regress $\exp(r(\xb_T)/\alpha)$ on $\xb_0$. 

While the above procedure is mathematically sound, in practice, where $\alpha$ is small, it may face numerical instability.  We instead recommend the following alternative. Suppose $r(\xb_T)=k(\xb_0) + \epsilon$ where $\epsilon$ is noise under $\PP^{\pre}$. Then, 
\begin{align*}
v^{\star}_0(x)=k(x) + \alpha \log \EE_{\PP^{\pre}}[\exp(\epsilon/\alpha)|\xb_0=x]. 
\end{align*}
Therefore, we can directly regress $r(\xb_T)$ on $\xb_0$ since the difference between $k(x)$ and $v^{\star}_0(x)$ remains constant. The complete algorithm is in Algorithm~\ref{alg:awr}. For more advanced methods, refer to Appendix~\ref{sec:refined}.

\begin{algorithm}[!t]
\caption{Optimal Value Function Estimation }\label{alg:awr}
\begin{algorithmic}[1]
  \STATE {\bf Input}: Function class $\Acal \subset [\RR^d \to \RR]$
  \STATE  Generate a dataset $\Dcal$ that consists of pairs of $(x,y)$: $x\sim \nu_{\ini}$ and $y$ as $r(\xb_T)$ following the pre-trained SDE:
    \begin{align*} \textstyle 
    d \xb_t = f(t,\xb_t)dt +  \sigma(t)d\wb_t. 
\end{align*}
  \STATE  Run an empirical risk minimization: 
  \begin{align*}\textstyle 
     \hat a=\argmin_{a\in \Acal} \sum_{(x,y)\sim \Dcal} \{\hat a(x)- y  \}^2.
  \end{align*}
    \STATE {\bf Output}: $\hat a$
\end{algorithmic}
\end{algorithm}

\vspace{-2mm}
\subsection{Sources of Approximation Errors} 
\vspace{-2mm}
Lastly, we explain the factors contributing to approximation errors in our algorithm. First, our method relies on the precision of neural SDE solvers, specifically, the expressiveness of neural networks and errors from discretization \citep{tzen2019theoretical}. Similarly, in the sampling phase, we also incur errors stemming from discretization. Additionally, our method relies on the expressiveness of another neural network in value function estimation.

\vspace{-3mm}
\section{Experiments}
\vspace{-2mm}

{We compare \agl against several baselines across two domains. Our goal is to check that \agl enables us to obtain diffusion models that generate high-reward samples while avoiding reward collapse and preserving diversity. We will begin by providing an overview of the baselines, describing the experimental setups, and specifying the evaluation metrics employed across all three domains. For more detailed information on each experiment, including dataset, architecture, and hyperparameters, refer to Appendix~\ref{sec:experiment}. 

\vspace{-2mm}
\paragraph{Methods to compare.}  We compare the following:

\setlist{nolistsep}
\begin{itemize} 
    \item \textbf{\agl:} Our method. 
    \item \textbf{NO KL:} This is \agl without the KL regularization and initial distribution learning. This essentially corresponds to AlignProp~\citep{prabhudesai2023aligning} and DRaFT~\citep{clark2023directly} in the discrete-time formulation. While several ways to mitigate reward collapse in these papers are discussed, we will compare them with our work later in Section~\ref{subsec:bio_oracle}. 
    \item \textbf{PPO + KL:} KL-penalized RL finetuning with PPO \citep{schulman2017proximal}. This is an improved version of both DPOK \citep{fan2023dpok} and DDPO \citep{black2023training} with a KL penalty applied to the rewards. For details, refer to Appendix~\ref{subsec:baseline}. 
    \item \textbf{Guidance}: We train a reward model to predict the reward value $y$ from a sample $x$. We use this model to guide the sampling process \citep{dhariwal2021diffusion,graikos2022diffusion} toward high rewards.
    For details, refer to Appendix~\pref{subsec:baseline}. 
\end{itemize}

\vspace{-3mm}
\paragraph{Experiment setup.} In all scenarios, we start by preparing a diffusion model with a standard dataset, containing a mix of high- and low-reward samples. Then, we create a (nominal) reward function 
$r$ by training a neural network reward model on a dataset with reward labels $\{x^{(i)}, r^{\star}(x^{(i)})\}_{i=1}^n$, ensuring that $r$ closely approximates the ``genuine'' reward function $r^{\star}$ on the data distribution of the dataset (i.e., on the support of pre-trained diffusion model). Following existing work, we mostly evaluate performance in terms of $r$, and measure reward collapse using separate divergence and diversity metrics (except for one task, where $r^{\star}$ is known).

\vspace{-3mm}
\paragraph{Evaluation.} We record the reward $r(x_T)$ ((b1) in Equation~\eqref{eq:obje}), the KL divergence term ((b2) in Equation~\eqref{eq:obje}), and a measure of diversity among generated samples, defined as $\EE_{x\sim p,z\sim p}[d(x,z)]$, where $d(x,z)$ represents the distance between $x$ and $z$. %
In our results, we present the mean values of the reward (Reward), the KL term (KL-Div), and the diversity term (Div) across several trials.  

Our aim is to fine-tune diffusion models so that they have high (Reward), low (KL-div), and high (Div): that is, to produce diverse yet high reward samples from a distribution that stays close to the data. For one of our evaluation tasks (Section~\ref{subsec:bio_oracle}), we know the true function $r^\star$ (though it is \emph{not} provided to our algorithm), and therefore can directly measure the degree to which our method mitigates overoptimization by comparing the values of $r$ and $r^\star$ for our method and baselines. However, for the other tasks, we must rely on the (KL-div) and (Div) metrics, as well as qualitative examples, to evaluate this.

\vspace{-2mm}
\subsection{Protein and DNA Sequences}\label{subsec:bio}
\vspace{-1mm}

We study two distinct biological sequence tasks: GFP and TFBind \citep{trabucco2022design}.  In the GFP task, $x$ represents green fluorescent protein sequences, each with a length of 237, and $r^{\star}(x)$ signifies their fluorescence value \citep{sarkisyan2016local}. In the TFBind task, $x$ represents DNA sequences, each having a length of 8, while $r^{\star}(x)$ corresponds to their binding activity with human transcription factors \citep{barrera2016survey}.
Using these datasets, we proceed to train transformer-based diffusion models and oracles (details in Appendix~\ref{subsec:detaile_bio}). For the distance metric $d$, following \citet{ghari2023generative}, we use Levenshtein distance.

\vspace{-3mm}
\paragraph{Results.}

To start, we present the performance of the methods in Tables~\ref{tab:gfp} and \ref{tab:tfbind}. It's clear that \agl surpasses \textbf{PPO + KL} and \textbf{Guidance} in terms of rewards, maintains a smaller KL term, and achieves higher diversity scores. It's worth noting that while there is typically a tradeoff between the reward and KL term, even when fine-tuned diffusion models yield similar rewards, their KL divergences can vary significantly. This implies that, compared to \textbf{PPO + KL}, \agl effectively minimizes the KL term while maintaining high rewards. This reduced KL term translates to the alleviation of overoptimization, as in Section~\ref{subsec:bio_oracle}.

Secondly, we observe that entropy regularization plays a crucial role in generating diverse samples. Without entropy regularization, \agl tends to produce samples with high rewards (in terms of $r(x)$), but the samples are very similar.
The introduction of entropy regularization effectively mitigates this collapse problem. To illustrate this, we present histograms of fine-tuned samplers with varying $\alpha$ in Figure~\ref{fig:histogram}. It's noticeable that as $\alpha$ increases, the average reward degrades slightly, but we are able to obtain more diverse samples.

\begin{table}[]
    \centering
        \caption{Results for GFP. We set $\alpha = 0.1$ for \agl and \textbf{PPO + KL}. We normalize florence scores. Note the pre-trained model has $0.90$ (reward), $0.0$ (KL-div), and $2.4$ (div).   }
    \begin{tabular}{cccc}
    \toprule 
         &  Reward $(r)$ $\uparrow$ & KL-Div $\downarrow$ & Div $\uparrow$ \\ \midrule 
        \textbf{Guidance} & $0.94 \pm 0.01$   & $624$ & $2.1$ \\ 
   \textbf{PPO + KL}  &  $0.96 \pm 0.01$  &   $95$     &   $2.0$ \\ \hline 
   \rowcolor{Gray}
        \textbf{\agl} (Ours)  &  $\mathbf{0.98}\pm 0.00$    & $\bm{32}$  & $\bm{2.2}$ \\ 
   \bottomrule 
    \end{tabular} 
    \label{tab:gfp}
    \vspace{-2mm}
\end{table} 

\begin{table}[]
    \centering
        \caption{Results for TFBind. We set $\alpha =0.005 $ for \agl and \textbf{PPO + KL}. Note the pre-trained model has $0.45$ (reward), $0.0$ (KL-div), $5.3$ (div). }
    \begin{tabular}{cccc } \toprule 
         &  Reward  $(r)$ $\uparrow$ & KL-Div $\downarrow$ & Div $\uparrow$  \\ \midrule  
           \textbf{Guidance} & $0.81 \pm 0.03$   & $709$  &  $\bm{2.5}$ \\ 
   \textbf{PPO + KL}  &  $\mathbf{0.98}\pm 0.00$     &  $110$  & $2.2$  \\  \hline 
     \rowcolor{Gray}
      \textbf{\agl} (Ours)  &    $\mathbf{0.98}\pm 0.00$   & $\bm{82}$ & $\bm{2.5}$ \\ 
   \bottomrule 
    \end{tabular} 
        \vspace{-2mm}
    \label{tab:tfbind}
\end{table}

\begin{figure}[!t]
	\centering
 	\begin{subfigure}{.32\linewidth}
\includegraphics[width=\linewidth]{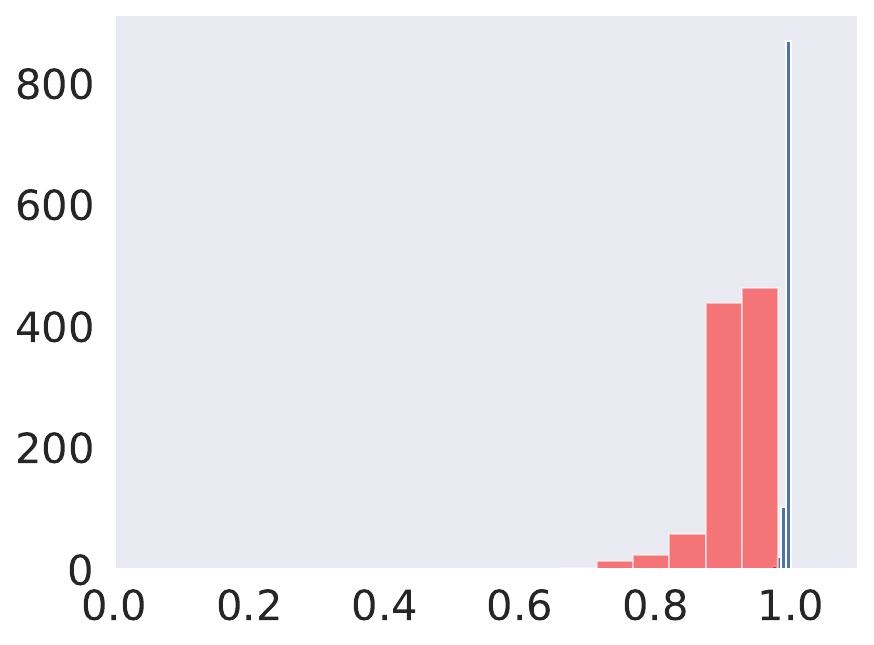}
		\caption{{\tiny \agl(5e-3)}}
	\end{subfigure}
	\begin{subfigure}{.33\linewidth}
\includegraphics[width=\linewidth]{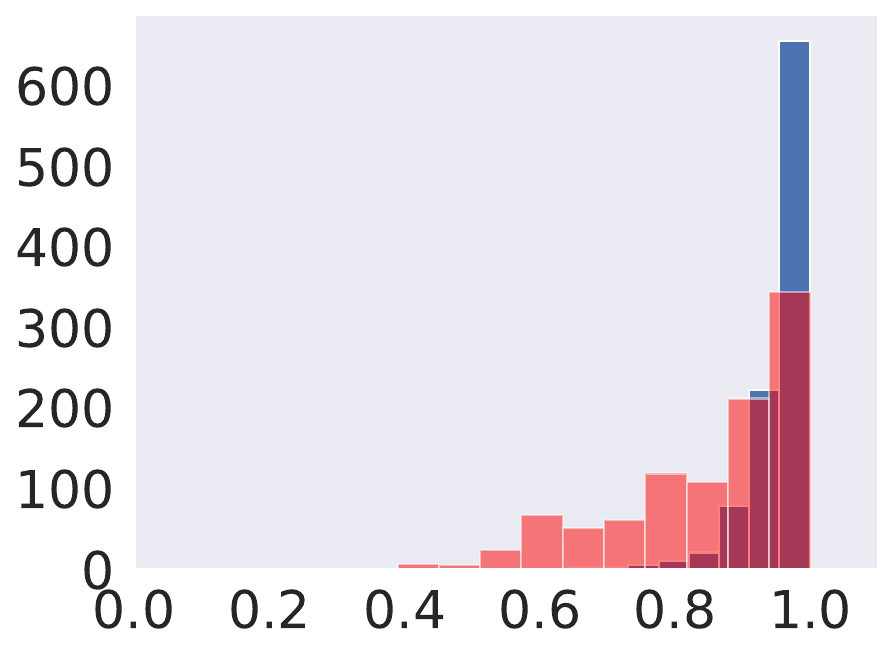}
		\caption{{\tiny \agl(e-2)}}
	\end{subfigure}
	\begin{subfigure}{.32\linewidth}
\includegraphics[width=\linewidth]{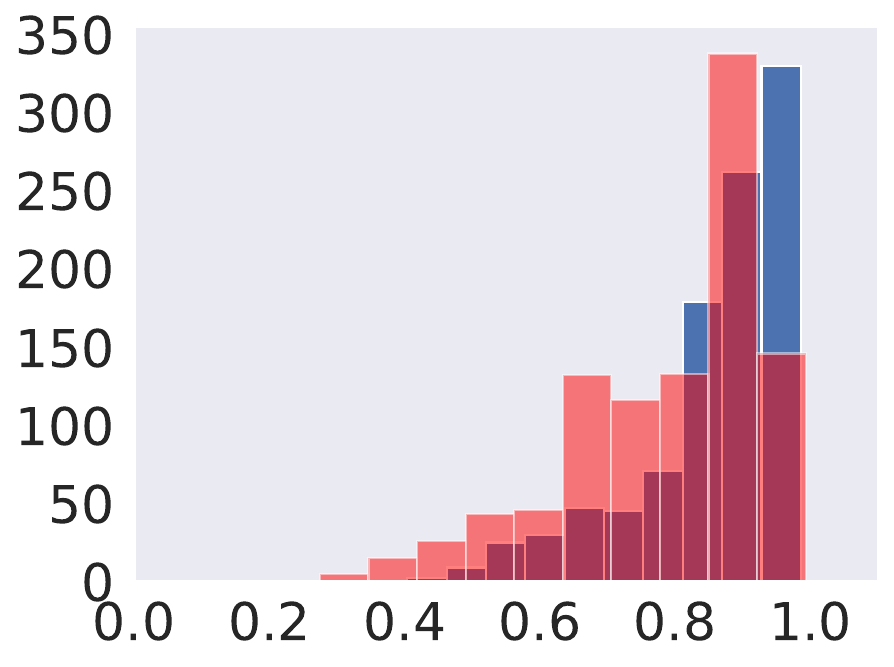}
		\caption{\tiny \agl(5e-2)}
	\end{subfigure}
  \begin{subfigure}{.32\linewidth}
\includegraphics[width=\textwidth]{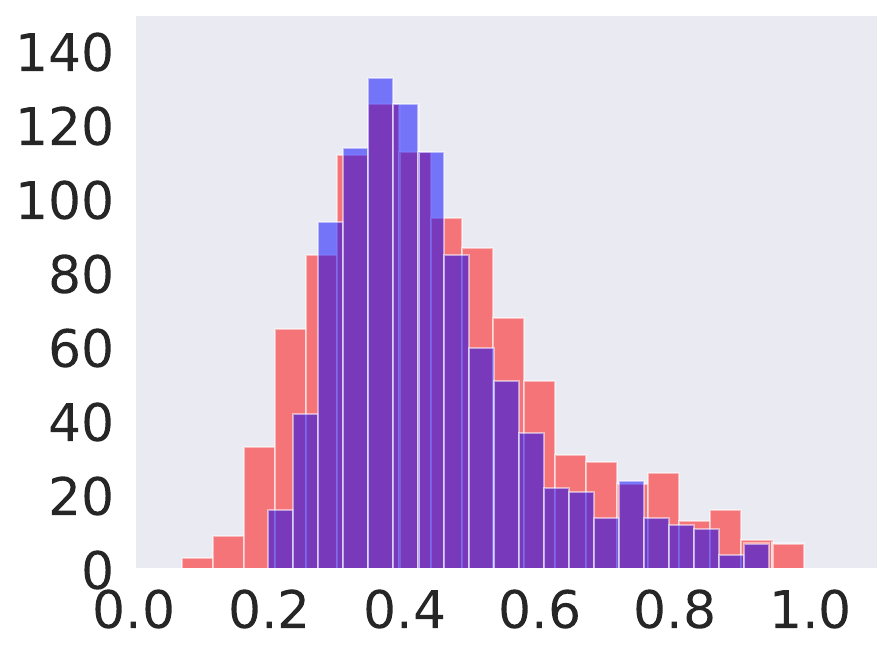}
	\caption{Pre-trained model}
	\end{subfigure}
   \begin{subfigure}{.32\linewidth}
\includegraphics[width=\textwidth]{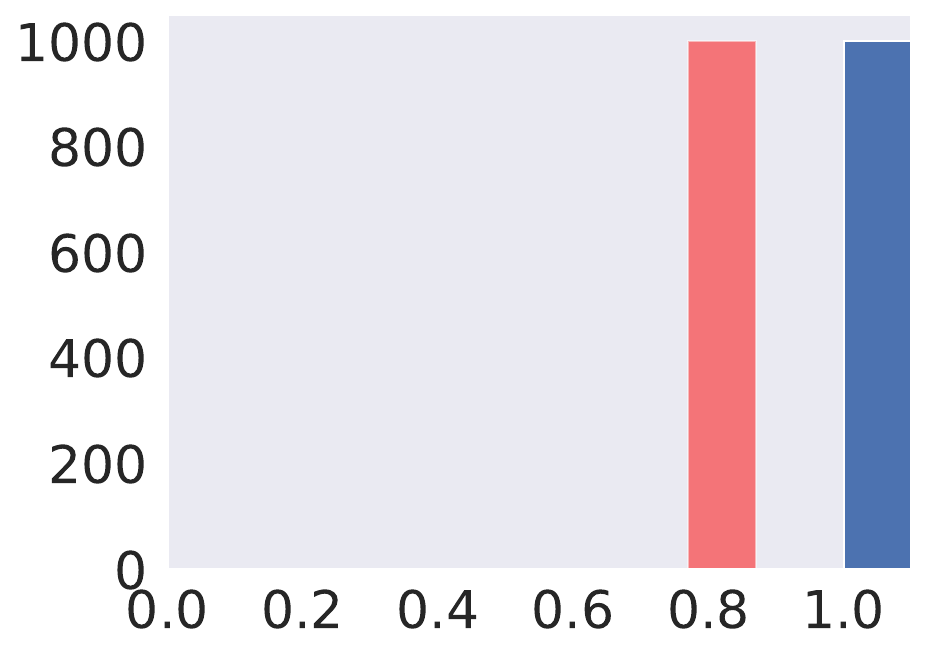}
	\caption{\textbf{NO KL}}
	\end{subfigure}
	\caption{Histograms of $1000$ samples generated by fine-tuned diffusions for TFBind in terms of \textcolor{red}{$r^{\star}(x)$} in \textcolor{red}{Red} and \textcolor{blue}{$r(x)$} in \textcolor{blue}{Blue}. In \textbf{No KL}, the same sample is generated (Div =0.0), and it suffers from overoptimization. \agl can effectively achieve both high $r$ and $r^{\star}$. The enlarged figure (a) is in Appendix~\ref{subsec:additional_image}.    } 
 \label{fig:histogram}
\end{figure}

\begin{table}[!t]
    \centering
        \caption{TFBind. We set $\alpha=0.01$ for \agl and PPO. For \textbf{Truncation}, we set $K=0.8T$. It is seen that \agl can circumvent overoptimization while other methods suffer from it. 
        }
    \begin{tabular}{ccc} \toprule 
         &  Reward $(r)$ $\uparrow$  & Reward $(r^{\star})$  $\uparrow$ \\ \midrule 
  \textbf{NO KL} & $ \mathbf{1.0}\pm 0.0$  & $0.76 \pm 0.02 $ \\
  \textbf{Guidance} &  $0.81\pm 0.03$   & $0.76 \pm 0.03 $  \\ 
 \textbf{PPO + KL} &   $0.987\pm 0.001 $  &   $0.84 \pm 0.01 $\\  \hline 
  \textbf{Random} &  $\mathbf{1.0}\pm 0.0$  & $0.77 \pm 0.01$ \\
  \textbf{Truncation} &  $\mathbf{1.0}\pm 0.0$  &   $0.78 \pm 0.01 $\\ \hline
    \rowcolor{Gray}  \agl  (Ours)  &   $0.989 \pm 0.001$   &  \textcolor{red}{$\mathbf{0.88} \pm 0.01$}\\  
   \bottomrule 
    \end{tabular} 
\label{tab:true_reward}
\end{table}

\vspace{-2mm}
\subsection{Quantitative Evaluation of Overoptimization}\label{subsec:bio_oracle}
\vspace{-2mm}

In TFBind, where we have knowledge of the genuine reward $r^{\star}$, we conduct a comparison between \agl and several baselines, presented in Table~\ref{tab:true_reward}. It becomes evident that the version without KL regularization achieves high values for $r$, but not for the true reward value $r^{\star}$. In contrast, our method can overcome overoptimization by effectively minimizing the KL divergence.

In addition, we compare algorithms that focus solely on maximizing $r(x_T)$ (i.e., ((b1) in Equation~\eqref{eq:obje}) with several techniques. For instance, the approach presented in \citet{clark2023directly} (DRaFT-K) can be adapted to our context by updating drift terms only in the interval $[K,T]$ rather than over the entire interval $[0,T]$ (referred to as \textbf{Truncation}). Similarly, AlignProp, as proposed by \citet{prabhudesai2023aligning}, can be applied by randomly selecting the value of $K$ at each epoch (referred to as \textbf{Random}).
However, in the case of TfBind, it becomes evident that these techniques cannot mitigate overoptimization.

\vspace{-2mm}
\subsection{Images}
\vspace{-2mm}

\begin{figure}[!t]
    \centering
\includegraphics[width=\linewidth]{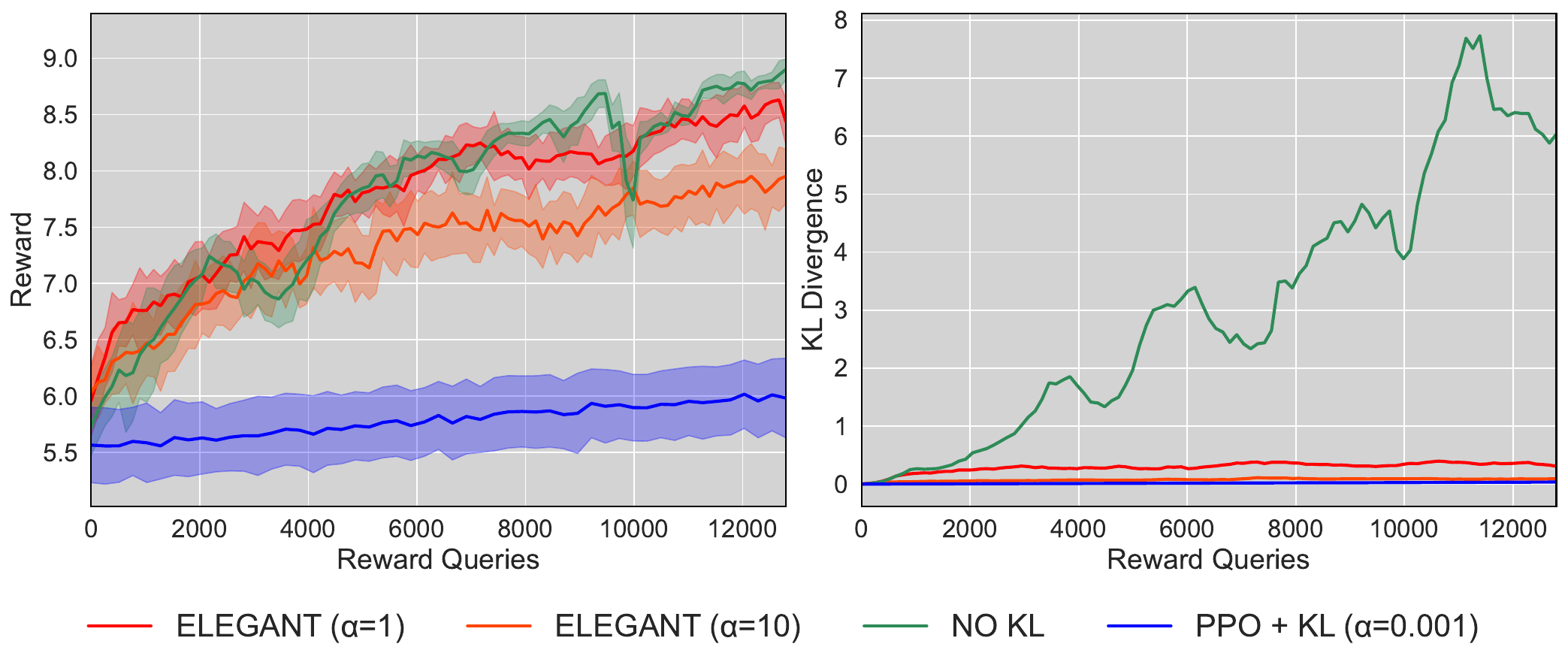}
\caption{Training curves of reward (left) and KL divergence (right) for fine-tuning aesthetic scores. The $x$ axis corresponds to the number of reward queries in the fine-tuning process. It shows that \agl\,is much faster than PPO, and \agl can effectively lower the KL divergence. } 
    \label{fig:training_curve}
\end{figure}

\begin{figure}[!t]
    \centering
    \includegraphics[width=0.7\linewidth]{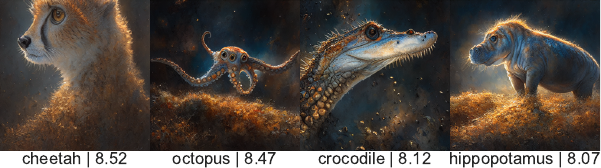}
    \caption{
    Diverse generated images by $\agl$ with high aesthetic scores.
    }
    \label{fig:enter-label}
\end{figure}

We use Stable Diffusion v1.5 as the pre-trained diffusion model~\citep{Rombach_2022_CVPR}. 
For $r(x)$, following existing works \citep{black2023training,prabhudesai2023aligning}, we use the LAION Aesthetics Predictor V2~\citep{schuhmann2022laion}, which is implemented as a linear MLP model on top of the OpenAI CLIP embeddings~\citep{radford2021clip}.

\vspace{-2mm}
\paragraph{Results.} 
\vspace{-2mm}

We plot the training curves of \agl and baselines in Figure~\ref{fig:training_curve}. First, it can be seen that the training speed of \agl is significantly faster than that of \textbf{PPO + KL}. Secondly, compared to \textbf{NO KL}, the KL divergence curves reveal that \agl maintains a consistently lower KL divergence throughout fine-tuning, indicating its ability to generate diverse samples without diverging significantly from the original data distribution. Some generated examples are shown in Figure~\ref{fig:enter-label}. We defer a more detailed comparison (including guidance) to Appendix~\ref{subsec:additional_image}. 

}

\vspace{-3mm}
\section{Conclusion}
\vspace{-2mm}

We propose a theoretically and empirically grounded, computationally efficient approach for fine-tuning diffusion models. This approach helps alleviate overoptimization issues and facilitates the generation of diverse samples. In future work, we plan to investigate the fine-tuning of recent diffusion models more tailored for biological or chemical applications \citep{watson2023novo,avdeyev2023dirichlet,gruver2023protein}.

\section*{Broader Impact}

This paper presents work whose goal is to advance the field of machine learning. There are many potential societal consequences of our work, none of which we feel must be specifically highlighted here.

\bibliographystyle{chicago}
\bibliography{rl}

\appendix 

\newpage 

\onecolumn 

\section{Additional Related Works}\label{sec:related}

In this section, we discuss additional related works.

\paragraph{Fine-tuning large language models.}
Much of the recent work in fine-tuning diffusion models is inspired by the wide success of fine-tuning large language models for various objectives such as instruction-following, summarization, or safety \citep{ouyang2022training,stiennon2020learning,bai2022constitutional}. Many techniques have been proposed to mitigate reward collapse in this domain, but KL regularization is the most commonly used \citep{gao2023scaling}. For a more comprehensive review, we direct readers to \citet{casper2023open}. In our experiments, we compare to a KL-penalized RL baseline, which is analogous to the current dominant approach in language model fine-tuning.

\paragraph{Sampling and control.}

Control-based approaches have been extensively employed for generating samples from known unnormalized probability densities in various ways \citep{tzen2019theoretical,bernton2019schr,heng2020controlled,zhang2021path,berner2022optimal,lahlou2023theory,zhang2023diffusion,bengio2023gflownet}.
Notably, the most pertinent literature relates to path integral sampling \citep{zhang2021path}. Nevertheless, our work differs in terms of our target distribution and focus, which is primarily centered on fine-tuning. Additionally, their method often necessitates assuming the initial distribution to be a Dirac delta distribution, which is an assumption we cannot make. Our research also shares connections with path integral controls \citep{theodorou2012relative,theodorou2010generalized, kappen2007introduction} and the concept of control as inference \citep{levine2018reinforcement}. However, our focus lies on the diffusion model, while their focus lies on standard RL problems.

\paragraph{Markov Chain Monte Carlo (MCMC).} MCMC-based algorithms are commonly used for sampling from unnormalized densities that follow a proportionality of $\exp(r(x)/\alpha)$ \citep{girolami2011riemann,ma2019sampling}. Numerous MCMC methods have emerged, including the first-order technique referred to as MALA. The approach most closely related to incorporating MALA for fine-tuning is classifier-based guidance, as proposed in \citet{dhariwal2021diffusion,graikos2022diffusion}. However, implementing classifier-based guidance is known to be unstable in practice due to the necessity of training numerous classifiers \citep{clark2023directly}.

\section{Connection with Classifier Guidance}\label{ape:classfier}

Recall the optimal SDE is written as 
\begin{align*}
    d x_t = \{f(t,x_t) + u^{\star}(t, x_t)\}dt + \sigma(t) dw_t 
\end{align*}
where 
\begin{align*}
    u^{\star}(t, x_t) =  \sigma^2(t)  \nabla_{x} \left\{ \log \EE_{\PP^{\pre}}\left [\exp\left (\frac{r(\xb_T)}{\alpha} \right)|\xb_t=x \right]\right\}. 
\end{align*}
When $r(x_T)=k(x_t) + \epsilon_t$ for some noise $\epsilon_t$ and a function $k:\RR^d \to \RR$, as we discuss in Section~\ref{subsec:value}, the above is approximated as 
\begin{align}\label{eq:classfier_based}
    u^{\star}(t, x_t) =  \frac{\sigma^2(t)}{\alpha}  \nabla_{x}  \EE_{\PP^{\pre}}\left [ r(\xb_T)|\xb_t=x \right] . 
\end{align}
Hence, at inference time, this boils down to using 
\begin{align*}
      d x_t = \left \{f(t,x_t) +  \frac{\sigma^2(t)}{\alpha}  \nabla_{x_t}  \EE_{\PP^{\pre}}\left [ r(\xb_T)|\xb_t \right]
      \right \} + \sigma(t) dw_t. 
\end{align*}

This formulation has a close connection with classifier-based guidance. The naive adaptation of classifier guidance \citep{dhariwal2021diffusion} in our context is 
\begin{align*}
      d x_t = \{f(t,x_t) + s'(t,x_t)\}dt + \sigma(t) dw_t
\end{align*}
where 
\begin{align*}
    s'(t,x) = \nabla_x \log p(r(x_T)|x_t). 
\end{align*}
Supposing that $r(x_T)=k(x_t)+\epsilon$ where $\epsilon$ is a Gaussian noise with variance $\sigma$, this is equal to 
\begin{align*}
     \frac{1}{\sigma^2}\nabla_x \EE_{\PP^{\data}}[r(x_T)|x_t=x]. 
\end{align*}
Up to a constant, this takes the same form in \eqref{eq:classfier_based}.

\section{Proofs}\label{sec:proof}

\subsection{Intuitive proof of Theorem~\ref{thm:important} }

We first give an intuitive proof of Theorem~\ref{thm:important}. 

Let $\PP^{u}_{\cdot|0}(\cdot|x_0)$ be the induced distribution by the SDE:
\begin{align*}
    dx_t = \{f(t,x_t) + u(t,x_t)\}dt + \sigma(t) d w_t. 
\end{align*}
over $\Ccal$ conditioning on $x_0$. Similarly, let  $\PP^{\data}_{\cdot|0}(\cdot|x_0)$ be the induced distribution by the SDE:
\begin{align*}
        dx_t = f(t,x_t) dt + \sigma(t) d w_t 
\end{align*}
over $\Ccal$ conditioning on $x_0$.

Now, we calculate the KL divergence of $\PP^{\data}_{\cdot|0}(\cdot|x_0)$ and $\PP^{u}_{\cdot|0}(\cdot|x_0)$ . This is equal to 
\begin{align}\label{eq:KL}
     \KL(\PP^{u}_{\cdot|0}(\cdot|x_0) \| \PP^{\pre}_{\cdot|0}(\cdot|x_0) ) = \EE_{\{x_t\} \sim \PP^{u}_{\cdot|0}(\cdot|x_0) }\left[ \int_{0}^T \frac{1}{2} \frac{\|u(t, x_t)\|^2}{\sigma^2(t)  }dt \right]. 
\end{align}
This is because 
\begin{align*}
     \KL(\PP^{u}_{\cdot|0}(\cdot|x_0) \| \PP^{\pre}_{\cdot|0}(\cdot|x_0) ) &=\EE_{\PP^{u}_{\cdot|0}(\cdot|x_0)}\left[\frac{d \PP^{u}_{\cdot|0}(\cdot|x_0)}{d \PP^{\pre}_{\cdot|0}(\cdot|x_0)} \right]  \\
    &=\EE_{\PP^{u}_{\cdot|0}(\cdot|x_0)}\left[\int_{0}^T \frac{1}{2} \frac{\|u(t, x_t)\|^2}{\sigma^2(t)  }dt + \int_{0}^T  u(t,x_t) d w_t \right ] \tag{Girsanov theorem} \\ 
    &=\EE_{\PP^{u}_{\cdot|0}(\cdot|x_0) }\left[ \int_{0}^T \frac{1}{2} \frac{\|u(t, x_t)\|^2}{\sigma^2(t)  } dt \right].  \tag{Martingale property of Itô integral}
\end{align*}

Therefore, the objective function in \eqref{eq:obje} is equivalent to 
\begin{align}
   \mathrm{obj} = \EE_{\PP^{u,\nu}}[r(x_T)] - \alpha \KL( \PP^{u,\nu} \| \PP^{\pre} ). 
\end{align}
This is because
\begin{align*}
     & \EE_{\PP^{u,\nu}}[r(x_T)]  -  \alpha \KL(\nu \| \nu_{\ini} ) - \alpha \EE_{\PP^{u,\nu} }\left[ \int_{0}^T \frac{1}{2} \frac{\|u(t, x_t)\|^2}{\sigma^2(t)  } dt \right] \\ 
     & = \EE_{\PP^{u,\nu}}[r(x_T)] - \alpha \KL(\nu \| \nu_{\ini} )- \alpha \EE_{x_0 \sim \nu}\left [\KL(\PP^{u}_{\cdot|0}(\cdot|x_0) \| \PP^{\pre}_{\cdot|0}(\cdot|x_0) ) \right ]  \\
     & =    \EE_{\PP^{u,\nu}}[r(x_T)] - \alpha \KL( \PP^{u,\nu} \| \PP^{\pre} ).
\end{align*}
The objective function is further changed as follows: 
\begin{align*}
     \mathrm{obj} &= \EE_{\PP^{u,\nu}}[r(x_T)] - \alpha \KL( \PP^{u,\nu} \| \PP^{\pre} ) \\ 
     &= \underbrace{\EE_{x_T \sim \PP^{u,\nu}_T}[r(x_T)] - \alpha  \KL( \PP^{u,\nu}_T \| \PP^{\pre}_T )}_{\textbf{Term (a)}} - \underbrace{\alpha \EE_{x_T \sim \PP^{u,\nu}_T }[\KL( \PP^{u,\nu}_T(\tau|x_T) \| \PP^{\pre}_T(\tau|x_T) )]}_{\textbf{Term (b)}}   \}. 
\end{align*}
By optimizing (a) and (b) over $\PP^{u,\nu}$, we get 
\begin{align}
    \PP^{\star}_T(x_T) &= \exp(r(x_T)/\alpha)\PP^{\pre}(x_T)/C, \nonumber \\
    \PP^{\star}_T(\tau|x_T)  & = \PP^{\data}_T(\tau|x_T).  \label{eq:bridge_preserving}
\end{align}
Hence, we have 
\begin{align*}
     \PP^{\star}(\tau) =  \PP^{\star}_T(x_T) \times  \PP^{\star}_T(\tau |x_T) = \frac{ \exp(r(x_T)/\alpha)\PP^{\pre}(\tau)}{C}. 
\end{align*}

\begin{remark}
Some readers might wonder in the part we optimize over $\PP_{f,\nu}$ rather than $f,\nu$. Indeed, this step would go through when we use non-Markovian drifts for $f$. While we use Markovian drift, this part still goes through because the optimal drift needs to be known as Markovian anyway. We choose to present this proof first because it can more clearly convey our message of bridge preserving property in \eqref{eq:bridge_preserving}. We will formalize it in \pref{sec:detailed2} and \pref{sec:detailed}. 
\end{remark}

\subsection{Formal Proof of \pref{thm:important}} 
\label{sec:detailed2}

Firstly, we aim to show that the optimal conditional distribution over $\Ccal$ on $x_0$ (i.e., $\PP^{u^{\star}}_{\cdot |0}(\tau|x_0)$) is equivalent to 
\begin{align*}
    \frac{\PP^{\pre}_{\cdot|0}(\tau|x_0) \exp(r(x_T)/\alpha)}{C(x_0) },\quad C(x_0):= \exp(v^{\star}_0(x)/\alpha). 
\end{align*}
To do that, we need to check that the above is a valid distribution first. This is indeed valid because the above is decomposed into 
\begin{align}\label{eq:valid}
    \underbrace{ \frac{\exp(r(x_T)/\alpha)\PP^{\data}(x_T|x_0)  }{C(x_0)}}_{(\alpha 1)} \times \underbrace{\PP^{\pre}_{\cdot|0}(\tau|x_0,x_T)}_{ (\alpha 2)},  
\end{align}
and both $(\alpha 1),(\alpha 2)$ are valid distributions. Especially, for the term $(\alpha 1)$, we can check as follows: 
\begin{align*}
    C(x_0) = \int \exp(r(x_T)/\alpha) 
 d\PP^{\pre}_{T|0}(x_T|x_0)) =\EE_{\PP^{\pre}_{\cdot|x_0}}[\exp(r(x_T)/\alpha)] =   \exp(v^{\star}_0(x)/\alpha). \tag{Use Lemma~\ref{lem:Feynman} }
\end{align*}
Now, after checking \eqref{eq:valid} is a valid distribution, we calculate the KL divergence: 
\begin{align*}
&\KL \left (\PP^{u^{\star}}_{\cdot |0}(\tau|x_0) \bigg{\|} \frac{\PP^{\pre}_{\cdot|0}(\tau|x_0) \exp(r(x_T)/\alpha)}{C(x_0) }  \right )\\ 
    & = \KL(\PP^{u^{\star}}_{\cdot |0}(\tau|x_0) \|\PP^{\pre}_{\cdot|0}(\tau|x_0))-\EE_{ \PP^{u^{\star}}_{\cdot|0}(\cdot|x_0) }\left[r(x_T)/\alpha - \log C(x_0) |x_0 \right ]\\
    &= \EE_{  \PP^{u^{\star}}_{\cdot|0}(\cdot|x_0) }\left[\left\{ \int_{0}^T \frac{1}{2} \frac{\|u^{\star}(t, x_t)\|^2}{\sigma^2(t)  }\right\}dt - r(x_T)/\alpha+ \log C(x_0)\mid x_0  \right] \tag{Use \eqref{eq:KL}}\\
    &= - v^{\star}_0(x_0)/\alpha + \log C(x_0). \tag{Definition of optimal value function}
\end{align*}
Therefore, 
\begin{align*}
    \KL \left (\PP^{u^{\star}}_{\cdot |0}(\tau|x_0) \| \frac{\PP^{\pre}_{\cdot|0}(\tau|x_0) \exp(r(x_T)/\alpha)}{C(x_0) }  \right )=- v^{\star}_0(x_0)/\alpha + \log C(x_0)=0. 
\end{align*}
Hence, 
\begin{align*}
    \PP^{u^{\star}}_{\cdot |0}(\tau|x_0) = \frac{\PP^{\pre}_{\cdot|0}(\tau|x_0) \exp(r(x_T)/\alpha)}{C(x_0) }. 
\end{align*}

Now, we aim to calculate an exact formulation of the optimal initial distribution. We just need to solve 
\begin{align*}
    \argmax_{\nu'}\int v^{\star}_0(x)\nu'(x)-\alpha  \KL(\nu'\|\nu_{\ini}). 
\end{align*}
The closed-form solution is 
\begin{align*}
    \exp(v^{\star}_0(x)/\alpha)\nu_{\ini}(x)/C 
\end{align*}
where $C:=\int \exp(v^{\star}_0(x)/\alpha)\nu_{\ini}(x)dx.$

Combining all together, we have been proved that the induced trajectory by the optimal control and the optimal initial distribution is 
\begin{align*}
    \PP^{u^{\star}}_{\cdot |0}(\tau|x_0)=\frac{\PP^{\pre}_{\cdot|0}(\tau|x_0) \exp(r(x_T)/\alpha)}{C(x_0)}, \quad \nu^{\star} \sim \frac{ C(x_0) \nu_{\ini}(x_0)}{C }. 
\end{align*}
Therefore, 
\begin{align*}
 \PP^{u^{\star},\nu^{\star}}(\tau) &= \PP^{u^{\star}}_{\cdot |0}(\tau|x_0)\nu^{\star} (x_0) = \frac{\PP^{\pre}_{\cdot|0}(\tau|x_0) \exp(r(x_T)/\alpha)}{C(x_0)} \times  \frac{ C(x_0) \nu_{\ini}(x_0)}{C } \\
    &= \frac{\PP^{\pre}_{\cdot|0}(\tau|x_0)  \exp(r(x_T)/\alpha)\nu_{\ini}(x_0)}{C} =  \frac{\PP^{\pre}(\tau)  \exp(r(x_T)/\alpha)}{C}. 
\end{align*}

\paragraph{Marginal distribution at $t$.}

Finally, consider the marginal distribution at $t$. By marginalizing before $t$, we get 
\begin{align*}
    \PP^{\pre}(\tau_{[t,T]})\times \exp(r(x_T)/\alpha)/C. 
\end{align*}
Next, by marginalizing after $t$, 
\begin{align*}
    \PP^{\pre}_t(x)/C\times \EE_{\PP_{\pre}}[ \exp(r(x_T)/\alpha) |x_t=x].
\end{align*}
Using Feynman–Kac formulation in Lemma~\ref{lem:Feynman}, this is equivalent to 
\begin{align*}
    \PP^{\pre}_t(x )\exp(v^{\star}_t(x )/\alpha)/C. 
\end{align*}

\paragraph{Marginal distribution at $T$.}

We marginalize before $T$. We have the following
\begin{align*}
    \PP^{\pre}_{T}(x )\exp(r(x )/\alpha)/C. 
\end{align*}

\subsection{Another Formal Proof of \pref{thm:important}} \label{sec:detailed}

First, noting the loss in \eqref{eq:obje} becomes
\begin{align*}
    \EE_{x_0\sim \nu}[v^{\star}_{0}(x_0) - \alpha \KL(\nu(x_0)/\nu_{\ini}(x_0) ) ], 
\end{align*}
by optimizing over $\nu \in \Delta(\Xcal)$, we can easily prove that the optimal initial distribution is  
\begin{align*}
    \exp\left(\frac{v^{\star}_{0}(x)}{\alpha} \right)\nu_{\ini}(x)/C. 
\end{align*}
Hereafter, our goal is to prove that the marginal distribution at $t$ (i.e., $\PP^{\star}_t$) is indeed $g_t(x)$ defined by 
\begin{align*}
     g_t(x):= \exp\left(\frac{v^{\star}_{t}(x)}{\alpha} \right)\PP^{\data}_t(x)/C
\end{align*}

Using \pref{lem:Feynman}, we can show that the SDE with the optimal drift term is 
 \begin{align*}
      dx_t = \left \{f(t,x) +  \frac{\sigma^2(t)}{\alpha}\nabla v^{\star}_t(x) \right\}dt + \sigma(t) d w_t .  
 \end{align*}
Then, what we need to prove is that the density $g_t \in \Delta(\RR^d)$ satisfies the Kolmogorov forward equation:
\begin{align}\label{eq:require}
    \frac{g_t(x)}{dt} + \sum_i \frac{d}{d x^{[i]}}\left[\left \{f^{[i]}(t,x) +  \frac{\sigma^2(t)}{\alpha}\nabla_{x^{[i]}} v^{\star}_t(x) \right\}   g_t(x)\right]- \frac{\sigma^2(t) }{2}\sum_{i} \frac{d^2 g_t(x)}{d x^{[i]}d x^{[i]} }=0 
\end{align}
where $f = [ f^{[1]},\cdots,f^{[d]}]^{\top}$.  
Indeed, this \eqref{eq:require} is proved as follows: 
\begin{align*}
    & \frac{dg_t(x)}{dt} + \sum_i \frac{d}{d x^{[i]}}\left[\left \{f^{[i]}(t,x) +  \frac{\sigma^2(t)}{\alpha}\nabla_{x^{[i]}} v^{\star}_t(x) \right\}   g_t(x)\right]- \frac{\sigma^2(t) }{2}\sum_{i} \frac{d^2 g_t(x)}{d x^{[i]}d x^{[i]} } \\
    &= \frac{1}{C}\exp\left(\frac{v^{\star}_{t}(x)}{\alpha} \right)\left\{ \frac{d \PP^{\data}_t(x)}{dt} +  \sum_i \nabla_{x^{[i]}}(\PP^{\data}_t(x)f^{[i]}(t,x)) -\frac{\sigma^2(t) }{2}\sum_{i} \frac{d^2 \PP^{\data}_t(x)}{d x^{[i]}d x^{[i]} }  \right \} \\
    & + \frac{1}{C}\PP^{\data}_t(x) \left\{ \frac{d \exp(v^{\star}_{t}(x)/\alpha) }{dt} +  \sum_i f^{[i]}(t,x) \nabla_{x^{[i]}}(\exp(v^{\star}_{t}(x)/\alpha) ) -\frac{\sigma^2(t) }{2}\sum_{i} \frac{d^2 \exp(v^{\star}_{t}(x)/\alpha)}{d x^{[i]}d x^{[i]} }  \right \}  \\
    & + \frac{1}{C}\PP^{\data}_t(x) \times \frac{\sigma^2(t) }{2}\sum_{i} \frac{d^2 \exp(v^{\star}_{t}(x)/\alpha)}{d x^{[i]}d x^{[i]} } \\
    &= 0 + 0. 
\end{align*}
Note in the final step, we use 
\begin{align*}
    \frac{d \PP^{\data}_t(x)}{dt} +  \sum_i \nabla_{x^{[i]}}(\PP^{\data}_t(x)f^{[i]}(t,x)) -\frac{\sigma^2(t) }{2}\sum_{i} \frac{d^2 \PP^{\data}_t(x)}{d x^{[i]}d x^{[i]} }  = 0,
\end{align*}
which is derived from the Kolmogorov forward equation, and the optimal value function satisfies the following 
\begin{align*}
    \frac{\sigma^2(t) }{2}\sum_{i} \frac{d^2 \exp(v^{\star}_t(x)/\alpha)}{d x^{[i]}d x^{[i]}} +f \cdot \nabla \exp(v^{\star}_t(x)/\alpha) +\frac{d \exp(v^{\star}_t(x)/\alpha)}{d t} = 0, 
\end{align*}
which will be shown in the proof of \pref{lem:Feynman} as in  \eqref{eq:optimal}. 

Hence, \pref{eq:require} is proved, and $g_t$ is $d\PP^{\star}_t/d\mu$.

\subsection{Proof of \pref{lem:Feynman}}

From the Hamilton–Jacobi–Bellman (HJB) equation, we have 
 \begin{align}\label{eq:HJB}
     &\max_{u} \left \{\frac{\sigma^2(t)}{2}\sum_{i} \frac{d^2 v^{\star}_t(x)}{d x^{[i]}d x^{[i]}} +\{f+u\} \cdot \nabla v^{\star}_t(x) +\frac{d v^{\star}_t(x)}{d t}  - \frac{\alpha \|u\|^2_2}{2\sigma^2(t)}  \right \} =0.
 \end{align}
 where $x^{[i]}$ is a $i$-th element in $x$.  Hence, by simple algebra, we can prove that the optimal control satisfies 
 \begin{align*}
      u^{\star}(t,x) = \frac{\sigma^2(t)}{\alpha}\nabla v^{\star}_t(x)  .  
 \end{align*}
By plugging the above into the HJB equation \eqref{eq:HJB}, we get 
  \begin{align}\label{eq:HJB2}
     &\frac{\sigma^2(t)}{2}\sum_{i} \frac{d^2 v^{\star}_t(x)}{d x^{[i]}d x^{[i]}} +f \cdot \nabla v^{\star}_t(x) +\frac{d v^{\star}_t(x)}{d t} +  \frac{\sigma^2(t) \| \nabla v^{\star}_t(x) \|^2_2}{2\alpha} =0,
 \end{align}
 which characterizes the optimal value function. Now, using \eqref{eq:HJB2}, we can show 
  \begin{align*}
     &\frac{\sigma^2(t) }{2}\sum_{i} \frac{d^2 \exp(v^{\star}_t(x)/\alpha)}{d x^{[i]}d x^{[i]}} +f \cdot \nabla \exp(v^{\star}_t(x)/\alpha) +\frac{d \exp(v^{\star}_t(x)/\alpha)}{d t} \\ 
     &=\exp \left (\frac{v^{\star}_t(x)}{\alpha} \right)\times \left \{ \frac{\sigma^2(t)}{2}\sum_{i} \frac{d^2 v^{\star}_t(x)}{d x^{[i]}d x^{[i]}} +f \cdot \nabla v^{\star}_t(x) +\frac{d v^{\star}_t(x)}{d t} +  \frac{\sigma^2(t) \| \nabla v^{\star}_t(x) \|^2_2}{2\alpha} \right\} \\
     &=0.
 \end{align*}
 Therefore, to summarize, we have 
 \begin{align}\label{eq:optimal}
     &\frac{\sigma^2(t) }{2}\sum_{i} \frac{d^2 \exp(v^{\star}_t(x)/\alpha)}{d x^{[i]}d x^{[i]}} +f \cdot \nabla \exp(v^{\star}_t(x)/\alpha) +\frac{d \exp(v^{\star}_t(x)/\alpha)}{d t} = 0, \\
     & v^{\star}_T(x)=r(x). 
 \end{align}
 Finally, by invoking the Feynman-Kac formula \citep{shreve2004stochastic}, we obtain the conclusion:  
  \begin{align*}
    \exp\left(\frac{v^{\star}_{t}(x)}{\alpha} \right)= \EE_{\PP^{\pre}}\left [\exp \left (\frac{r(\xb_T)}{\alpha } \right)|\xb_t=x \right].
\end{align*}

\subsection{Proof of \pref{lem:optimal_drift}}

Recall  $ u^{\star}(t,x)=\frac{\sigma^2(t)}{\alpha}\times \nabla_x v^{\star}_t(x)$ from the proof of \pref{lem:Feynman}. Then, we have 
 \begin{align*}
     u^{\star}(t,x)&=\frac{\sigma^2(t)}{\alpha}\times \nabla_x v^{\star}_t(x)=  \frac{\sigma^2(t)}{\alpha} \times \alpha \frac{\nabla_x \exp(v^{\star}_t(x)/\alpha )}{\exp(v^{\star}_t(x)/\alpha ) }\\
     &= \sigma^2(t) \times \frac{\nabla_{x} \EE_{\PP^{\pre}}[\exp(r(\xb_T)/\alpha )|\xb_t=x ]}{\EE_{\PP^{\pre}}[\exp(r(\xb_T)/\alpha )|\xb_t=x ]}. 
    \end{align*}

\subsection{Proof of \pref{lem:joint}}

Recall 
\begin{align*}
    \PP^{\star}(\tau)= \PP^{\pre}(\tau) \exp(r(x_T)/\alpha)/C
\end{align*}
By marginalizing before $s$, we get 
\begin{align*}
    \PP^{\star}_{[s,T]}(\tau_{[s,T]})\times \exp(r(x_T)/\alpha)/C
\end{align*}
Next, marginalizing after $t$, 
\begin{align*}
    & \PP^{\pre}_{[s,t]}(\tau_{[s,t]})\times \EE_{\PP^{\pre}}[\exp(r(x_T)/\alpha)|x_t]/C \\
    &=\PP^{\pre}_{[s,t]}(\tau_{[s,t]})\exp(v^{\star}_t(x_t)/\alpha)/C. 
\end{align*}
Finally, by marginalizing between $s$ and $t$, the joint distribution at $(s,t)$ is 
\begin{align*}
    \PP^{\pre}_{s,t}(x_s,x_t)\exp(v^{\star}_t(x_t)/\alpha)/C. 
\end{align*}

\paragraph{Forward conditional distribution.}

\begin{align*}
    \PP^{\star}_{t|s}(x_t|x_s) = \frac{\PP^{\pre}_{s,t}(x_s,x_t)\exp(v^{\star}_t(x_t)/\alpha)/C}{\PP^{\pre}_{s}(x_s)\exp(v^{\star}_s(x_s)/\alpha)/C }= \PP^{\pre}_{s|t}(x_t|x_s) \frac{\exp(v^{\star}_t(x_t)/\alpha)}{\exp(v^{\star}_s(x_s)/\alpha) }. 
\end{align*}

\paragraph{Backward conditional distribution.}

\begin{align*}
    \PP^{\star}_{s|t}(x_s|x_t) = \frac{\PP^{\pre}_{s,t}(x_s,x_t)\exp(v^{\star}_t(x_t)/\alpha)/C}{\PP^{\pre}_{s}(x_t)\exp(v^{\star}_t(x_t)/\alpha)/C }= \PP^{\pre}_{s|t}(x_s|x_t). 
\end{align*}

\subsection{Proof of \pref{lem:brdige}}

Use a statement in \eqref{eq:bridge_preserving} the proof of \pref{thm:important}. 

\subsection{Proof of \pref{thm:justification}}

We omit the proof since it is the same as the proof of \pref{thm:important}. 

\section{Diffusion models}

\label{subsec:continuous}

In this section, we provide an overview of continuous-time generative models. The objective is to train a SDE in such a way that the marginal distribution at time $T$ follows $p_{\pre}$. While $p_{\pre}$ is not known, we do have access to a dataset that follows this distribution. 

\paragraph{Denoising diffusion models} 

DDPMs \citep{song2020score} are a widely adopted class of generative models. We start by considering a forward stochastic differential equation (SDE) represented as:
\begin{align}\label{eq:forward}
    d\yb_t = -0.5 \yb_t d t +d\wb_t, \yb_0\sim p_{\pre}, 
\end{align}
defined on the time interval $[0,T]$. As $T$ tends toward infinity, the limit of this distribution converges to $\Ncal(0,\II_{d})$, where $\II_d$ denotes a $d$-dimensional identity matrix. Let $\QQ$ be a measure on $\Ccal$ induced by the forward SDE \eqref{eq:forward}. Consequently, a generative model can be defined using its time-reversal: ${\xb_t}={\yb_{T-t}}$, which is characterized by:
\begin{align*}
     d \xb_t = \{0.5 \xb_t-\nabla \log \QQ_{T-t}(\xb_t)\} d t + d \wb_t , \, \xb_0\sim \Ncal(0,\II_d).
\end{align*} 
A core aspect of DDPMs involves learning the score $\nabla \log \QQ_t$ by optimizing the following loss with respect to $S$:
\begin{align*}
 \EE_{\QQ}[\|\nabla \log \QQ_{t|0}(\xb_t|\xb_0)- S(t,\xb_t)\|^2_2 ]. 
\end{align*}
A potential limitation of the above approach is that the forward SDE \eqref{eq:forward} might not converge to a predefined prior distribution, such as $\Ncal(0,\II_d)$, within a finite time $T$. To address this concern, we can employ the Schrödinger Bridge formulation \citep{de2021diffusion}.

\paragraph{Diffusion Schrödinger Bridge.} 
A potential bottleneck of the diffusion model is that the forward SDE might not converge to a pre-specified prior $\Ncal(0,\II_d)$ with finite $T$. To mitigate this problem, \citet{de2021diffusion} proposed the following Diffusion Schrödinger Bridge. Being inspired by Schrödinger Bridge formulation \citep{schrodinger1931umkehrung} they formulate the problem: 
\begin{align*}
   \argmin_{\PP} \KL(\PP \| \PP_{\reef} )\,\mathrm{s.t.} \PP_0=\nu_{\ini},\PP_T=p_{\pre}. 
\end{align*} 
where $\PP_{\reef}$ is a reference distribution on $\Ccal$ such as a Wiener process, $\PP_0$ is a margin distribution of $\PP$ at time $0$, and $\PP_T$ is similarly defined. To solve this problem, \citet{de2021diffusion} proposed an iterative proportional fitting, which is a continuous extension of the Sinkhorn algorithm \citep{cuturi2013sinkhorn}, while learning the score functions from the data as in DDM. 

\paragraph{Bridge Matching.}

In DDPMs, we have formulated a generative model based on the time-reversal process and aimed to learn a score function $\nabla \log \QQ_{t}$ from the data. Another recent popular approach involves emulating the reference Brownian bridge given $0,T$ with fixed a pre-defined initial distribution $\nu_{\ini}$ and a data distribution $p_{\fun}$
at time $T$ \citep{shi2023diffusion,liu2022let}. To elaborate further, let's begin by introducing a reference SDE:
\begin{align*}
     d \bar \xb_t = \sigma(t) d \wb_t, \quad \bar \xb_0\sim \nu_{\ini}.
\end{align*}
Here, we overload the notation with $\QQ$ to indicate an induced SDE. The Brownian bridge $\QQ_{\cdot|[0,T]}(\cdot |x_0, x_T)$ is defined as:
\begin{align*}
    d \bar \xb^{0,T}_t = \sigma(t)^2 \nabla \log \QQ_{T|t}(x_T| \bar \xb^{0,T}_t) +\sigma(t) d\wb_t,  \xb^{0,T}_0=x_0, 
\end{align*}
where $\xb^{0,T}_T=x_T$. The explicit calculation of this Brownian bridge is given by:
\begin{align*}
     \nabla \log \QQ_{T|t}(x_T| \bar \xb^{0,T}_t)=\frac{x_T-\xb_t}{T-t}.  
\end{align*}
Now, we define a target SDE as follows:
\begin{align*}
     d \xb_t = f(t,\xb_t) d t +\sigma(t) d \wb_t, \quad \xb_0\sim \nu_{\ini}.
\end{align*}
This SDE aims to induce the same Brownian bridge as described earlier and should generate the distribution $p_{\pre}$ at time $T$. We use $\PP$ to denote the measure induced by this SDE. While the specific drift term $f$ is unknown, we can sample any time point $t (0<t<T)$ from $\PP$ by first sampling $x_0$ and $x_T$ from $\nu_{\ini}$ and $p_{\fun}$, respectively, and then sampling from the reference bridge, i.e., the Brownian bridge. To learn this drift term $f$, we can use the following characterization:
\begin{align*}
   f(t,\xb_t)=\EE_{\PP}[\nabla \log \QQ_{T|t}(\xb_T|\xb_t) |\xb_t=x_t].
\end{align*}
Subsequently, the desired drift term can be learned using the following loss function with respect to $f$:
\begin{align*}
    \EE_{\PP}[\| \nabla \log \QQ_{T|t}(\xb_T|\xb_t) - f(t,\xb_t) \|^2 ].
\end{align*}
Bridge matching can be formulated in various equivalent ways, and for additional details, we refer the readers to \citet{shi2023diffusion,liu2022flow}. 

\section{Details of Implementation}

\subsection{Implementation Details of Neural SDE}\label{subsec:neuralSDE}

We aim to solve 
\begin{align*}
    \argmax_{\bar f:[0,T]\times \Zcal \to \Zcal  }L( \zb_T),
d\zb_t = \bar f(t,\zb_t) dt + \bar g(t) d\wb_t, \zb_0 \sim \bar \nu. 
\end{align*}

Here is a simple method we use. Regarding more details, refer to \citet{kidger2021efficient,chen2018neural}. 

Suppose that $\bar f(\cdot;\theta)$ is parametrized by $\theta$. Then, we update this $\theta$ with SGD. Consider at iteration $j$. Fix $\theta_j$ in $\bar f(t,\zb_t;\theta_j)$. Then, by simulating 
an SDE with 
\begin{align*}\textstyle
   d\zb_t = \bar f(t,\zb_t;\theta) dt + \bar g(t) d\wb_t, \zb_0\sim \bar \nu,  
\end{align*}
we obtain $N$ trajectories
\begin{align*}\textstyle
    \{\zb^{(i)}_0,\cdots,\zb^{(i)}_T\}_{i=1}^N.  
\end{align*}
In this step, we are able to use any off-the-shelf discretization methods. For example, starting from $\zb^{(i)}_0\sim \nu$, we are able to obtain a trajectory as follows: 
\begin{align*}\textstyle
    \zb^{(i)}_t =\zb^{(i)}_{t-1} +  \bar f(t-1,\zb^{(i)}_{t-1};\theta)\Delta t + \bar g(t-1)\Delta w_t,\,\quad \Delta w_t\sim \Ncal(0, (\Delta t)^2). 
\end{align*}
Finally, using automatic differentiation, we update $\theta$ with the following: 
\begin{align*}
   \theta_{j+1}=\theta_j - \rho \nabla_{\theta}\left\{ \frac{1}{N}\sum_{i=1}^N L(\zb^{(i)}_T) \right \} \bigg{|}_{\theta=\theta_j} 
\end{align*}
where $\rho $ is a learning rate. We use Adam in this step for the practical selection of the learning rate $\rho$. 

\subsection{More Refined Methods to Learn Value Functions}\label{sec:refined}

We can directly get a pair of $x$ and $y$. Here, $\{x^{(i)}\} \sim \nu_{\ini}$ and 
\begin{align*}
     \hat y^{(i)}:= \alpha \log \hat \EE_{\PP^{\pre}}[\exp(r(\xb_T)/\alpha)|\xb_0=x^{(i)}] 
\end{align*}
where $\hat \cdot$ means Monte Carlo approximation for each $x_i$. In other words, 
\begin{align*}
    \hat \EE_{\PP^{\pre}}[\exp(r(\xb_T)/\alpha)|\xb_0=x_i]:= \frac{1}{n}\sum_{j=1}^n  \exp(r(\xb^{(i,j)}_T)/\alpha) 
\end{align*}
where $\{r(\xb^{(i,j)}_T\}$ is a set of samples following $\PP^{\pre}$ with initial condition: $\xb_0=x_i$. Then, we are able to learn $a$ using the following ERM: 
\begin{align*}
    \hat a =\argmin_{a \in \Acal} \sum_{i=1}^n \{a(x^{(i)})-\hat y^{(i)}\}^2. 
\end{align*}

\subsection{Baselines} \label{subsec:baseline}

We consider the following baselines. 

\paragraph{PPO + KL.}
Considering the discretized formulation of diffusion models  \citep{black2023training,fan2023dpok}, we use the following update rule:
\begin{align} \label{eq:natural}
  & \nabla_{\theta}\EE_{\Dcal}\sum_{t=1}^T \left [\min\left \{\tilde r_t(x_0,x_t)\frac{p(x_t|x_{t-1};\theta)}{p(x_t|x_{t-1};\theta_{\text{old}})}, \tilde r_t(x_0,x_t) \cdot \mathrm{Clip}\left (\frac{p(x_t|x_{t-1};\theta)}{p(x_t|x_{t-1};\theta_{\text{old}})}, 1-\epsilon,1+\epsilon\right ) \right\} \right],   \\
   & \tilde r_t(x_0,x_t) = - r(x_T) + \underbrace{\alpha \frac{\|u(t,x_t;\theta)\|^2 }{2 \sigma^2(t)}}_{\text{KL term}},\quad p(x_t|x_{t-1};\theta)=\Ncal(u(t,x_t;\theta)+f(t,x_t),\sigma(t)) \label{eq:natural_2}
\end{align}
where $f(t, x_t)$ is a pre-trained drift term and $\theta$ is a parameter to optimize. 

Note that DPOK~\citep{fan2023dpok} uses the following update: 
\begin{align*} 
  & \nabla_{\theta}\EE_{\Dcal}\sum_{t=1}^T \left [\min\left \{ -r(x_0)\frac{p(x_t|x_{t-1};\theta)}{p(x_t|x_{t-1};\theta_{\text{old}})},  - r(x_0) \cdot \mathrm{Clip}\left (\frac{p(x_t|x_{t-1};\theta)}{p(x_t|x_{t-1};\theta_{\text{old}})}, 1-\epsilon,1+\epsilon\right )\right\} +  \underbrace{\alpha \frac{\|u(t,x_t;\theta)\|^2 }{2\sigma^2(t)}}_{\text{KL term} } \right]
\end{align*}
where the KL term is directly differentiated. We did not use the DPOK update rule because DDPO appears to outperform DPOK even without a KL penalty (\citet{black2023training}, Appendix C), so we implemented this baseline by modifying the DDPO codebase to include the added KL penalty term (Equation~\eqref{eq:natural_2}).

\paragraph{Guidance.}

We use the following implementation of guidance \citep{dhariwal2021diffusion}: 
\begin{itemize}
\item For each $t \in [0,T]$, we train a model: $\PP_t(y|x_t)$ where $x_t$ is a random variable induced by the pre-trained diffusion model. 
    \item We fix a guidance level $\gamma \in \RR_{>0}$, target value $y_{\condition} \in \RR$, and at 
    inference time (during each sampling step), we use the following score function 
\begin{align*}
    \nabla_x \log{\PP_t(y = y_{\condition}|x)} = \nabla_x \log{\PP_t(x)} + \gamma \nabla_x \log{\PP_t(y=y_{\condition}|x)}. 
\end{align*}
\end{itemize}

A remaining question is how to model $p(y|x)$. In our case, for the biological example, we make a label depending on whether $x$ is top $10\%$ or not and train a binary classifier. In image experiments, we construct a Gaussian model: $p(y|x)=\mathcal{N}(y-\mu_\theta(x), \sigma^2)$ where $y$ is the reward label, $\mu_\theta$ is the reward model we need to train, and $\sigma$ is a fixed hyperparameter.

\section{Experiment Details}\label{sec:experiment}

\subsection{Details for Tasks in Biological Sequences}\label{subsec:detaile_bio}

\subsubsection{Dataset.}

\paragraph{TFBind8.} The number of original dataset size is $65792$. Each data consists of a DNA sequence with $8$-length. We represent each data as a one-hot encoding vector with dimension $8\times 4$. To construct diffusion models, we use all datasets. We use half of the dataset to construct a learned reward $r$ to make a scenario where oracles are imperfect. 

\paragraph{GFP.}  The original dataset contains $56,086$ data points, each comprising an amino acid sequence with a length of $237$. We represent each data point using one-hot encoding with a dimension of $237 \times 20$. Specifically, we model the difference between the original sequence and the baseline sequence. For our experiments, we selected the top $33,637$ samples following \citep{trabucco2022design} and trained diffusion models and oracles using this selected dataset.

\subsubsection{Structure of Neural Networks.}

We describe the implementation of neural networks in more detail. 

\paragraph{Diffusion models and fine-tuning.}

For diffusion models in TFBind, we use a neural network to model score functions in Table~\ref{tab:diffusion_tfbind}. We use a similar network for the GFP dataset and fine-tuning parts. 
\begin{table}[!t]
    \centering
        \caption{Architecture of diffusion models for TFBind }    \label{tab:diffusion_tfbind}
    \begin{tabular}{c c c c }
    \toprule 
     Layer    &  Input Dimension  & Output dimension  & Explanation \\ \hline 
      1   &    $1$ $(t)$ &   $256$ ($t'$)    & Get time feature  \\
      1   & $8\times 4$ $(x)$ &   $64$ ($x'$)     & Get positional encoder  (Denote $x'$)  \\ 
    2  &  $8\times 4 + 256 + 64$ $(x,t,x')$  &  $64$ $(\bar x)$  & Transformer encoder  \\
    3 & $ 64$  $(\bar x)$ &   $8 \times 4  $  $(x)$ &  Linear 
     \\ \bottomrule  
     \end{tabular}
\end{table}

\paragraph{Oracles to obtain score functions.} To construct oracles in TFBind, we employ the neural networks listed in Table~\ref{tab:oralce_TFBind}. For GFP, we utilize a similar network.

\begin{table}[!t]
    \centering
      \caption{Architecture of oracles for TFBind}
    \label{tab:oralce_TFBind}
    \begin{tabular}{cccc} \toprule 
         &  Input dimension & Output dimension &  Explanation \\  \hline 
      1   & $8\times 4 $   &   $500$  & Linear \\ 
      1   &  $500$ &  $500$  & ReLU  \\ 
      2  &  $500$ &  $200$   & Linear \\
     2    &  $200$ &  $200$  & ReLU  \\ 
    3  &  $200$ &  $1$   & Linear \\
     3    &  $200$ &  $1$  & ReLU  \\
     4  & $1$ & $1$  & Sigmoid \\ \bottomrule 
    \end{tabular}
\end{table}

\subsubsection{Hyperparameters} 

We report a set of important hyperparameters in Table~\ref{tab:hyper}. 

\begin{table}[!t]
    \centering
     \caption{Primary hyperparameters for fine-tuning. For all methods, we use the Adam optimizer. }
    \begin{tabular}{c|c|cc}\toprule 
   Method  &   Type     &  GFP  
&  TFBind   \\  \hline 
  \multirow{7}{*}{\agl}  &   Batch size  & $128$ & $128$ \\
   & Sampling for neural SDE &  \multicolumn{2}{c}{Euler Maruyama} \\ 
     & Step size (fine-tuning)  & $50$ & $50$  \\ 
     & Epochs (fine-tuning) & $50$ &  $50$ \\
     \hline 
 \textbf{PPO}   &  Batch size & $128$ & $128$ \\ 
 & $\epsilon$ & $0.1$ & $0.1$\\
  & Epochs & $100$ & $100$ \\  
  \hline  
 \textbf{Guidance}  & Guidance level  &  $30$ & $30$ \\ \hline  
 \textbf{Pre-trained diffusion}  & Forward SDE  & \multicolumn{2}{c}{Variance preserving} \\
 & Sampling way &  \multicolumn{2}{c}{Euler Maruyama}\\  
  \bottomrule  
    \end{tabular}
    \label{tab:hyper}
\end{table}

\subsubsection{Additional Results} \label{subsec:enlarge}

We add the enlarged figure of \agl (0.005) in Table~\ref{fig:histogram}. 

\begin{figure}[!t]
    \centering
    \includegraphics[width = 0.7\textwidth]{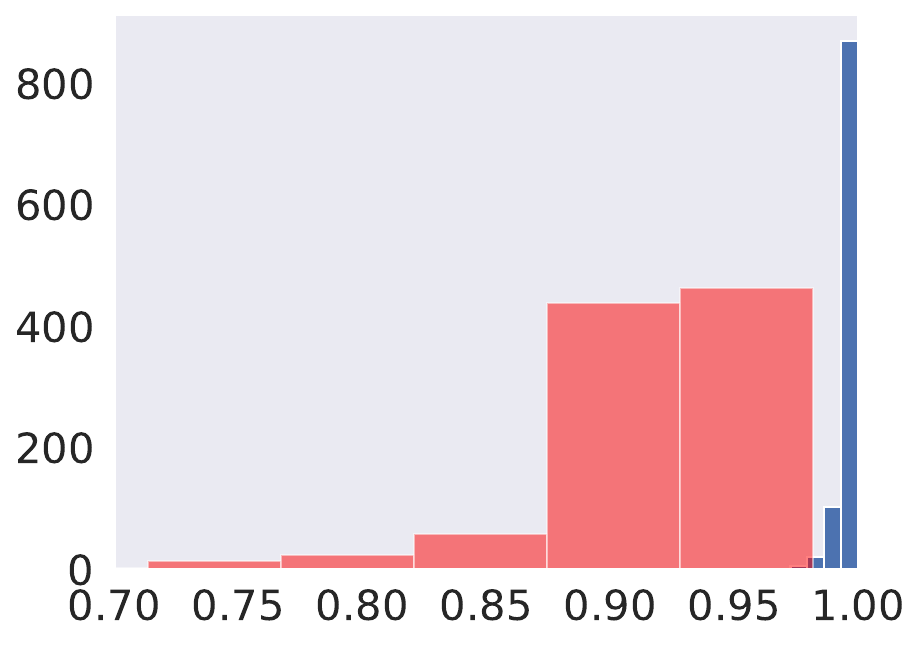}
    \caption{Enlarge histogram of \agl (0.005) in Table~\ref{fig:histogram}. }
    \label{fig:enlarged}
\end{figure}

\subsection{Details for Image Tasks}

Below, we explain the training details and list hyperparameters in Table~\ref{tab:hyper_params}. 

\subsubsection{Details of Implementation} 

We use 4 A100 GPUs for all the image tasks.
We use the AdamW optimizer~\citep{loshchilov2018decoupled} with $\beta_1=0.9, \beta_2=0.999$ and weight decay of $0.1$.
To ensure consistency with previous research, in fine-tuning, we also employ training prompts that are uniformly sampled from $50$ common animals~\citep{black2023training,prabhudesai2023aligning}.

\paragraph{Sampling.} We use the DDIM sampler with $50$ diffusion steps~\citep{song2020denoising}.
Since we need to back-propagate the gradient of rewards through both the sampling process producing the latent representation and the VAE decoder
used to obtain the image, memory becomes a bottleneck. We employ two designs to alleviate memory usage following ~\citet{clark2023directly,prabhudesai2023aligning}:
(1) Fine-tuning low-rank adapter (LoRA) modules~\citep{hu2021lora} instead of tuning the original diffusion
weights, and (2) Gradient checkpointing for computing partial derivatives on demand~\citep{gruslys2016memory, chen2016training}. The two designs make it possible to back-propagate gradients through all 50 diffusing steps in terms of hardware.

\begin{table}[!t]
\centering
\caption{Training hyperparameters.}
\begin{tabular}{lc}
\toprule
 Hyperparameter &\\ \hline
Learning rate& 1e-4 \\
Batch size per GPU& 4 \\
Clip grad norm & 5.0\\
\hline
DDIM steps & 50 \\
Guidance weight & 7.5 \\
Samples per iteration & 128 \\
\bottomrule
\end{tabular}
\label{tab:hyper_params}
\end{table}

\paragraph{Guidance.} To train the classifier, we use the AVA dataset~\citep{murray2012ava} which includes more than 250k evaluations (i.e., 20 times more samples than our $\agl$implementation, cf. Figure~\ref{fig:training_curve}). We implement the classifier (i.e., reward model) using an MLP model that takes the concatenation of sinusoidal time embeddings (for time $t$) and CLIP embeddings~\citep{radford2021clip} (for $x_t$) as input. The implementation is based on the public RCGDM~\citep{yuan2023reward} codebase\footnote{\url{https://github.com/Kaffaljidhmah2/RCGDM}}.

\subsubsection{Additional Results}\label{subsec:additional_image}

\begin{table}[!t]
    \centering
\caption{Results for aesthetic scores. We set $\alpha=1$ for \agl and set $\alpha=0.001$ for \textbf{PPO + KL}. $(\cdot)$ are confidence intervals.
}
    \begin{tabular}  {p{0.20\linewidth}p{0.16\linewidth}  p{0.10\linewidth} }    
    \toprule 
         &  Reward $(r)$ $\uparrow$ & KL-Div $\downarrow$  \\ \midrule 
         \textbf{NO KL} & $8.80 \pm 0.08$  & $6.03$ \\ 
        \textbf{Guidance} & $6.41 \pm 0.19$  & $5.99$ \\ 
   \textbf{PPO + KL}  &  $5.98 \pm 0.35$  &   $0.03$    \\ \hline 
  \rowcolor{Gray}\textbf{\agl} (Ours)  &  $8.62 \pm 0.16$    & $0.34$ \\ 
   \bottomrule 
    \end{tabular} 
    \label{tab:image}
\end{table}

\begin{figure}[!t]
    \centering
\includegraphics[width=\linewidth]{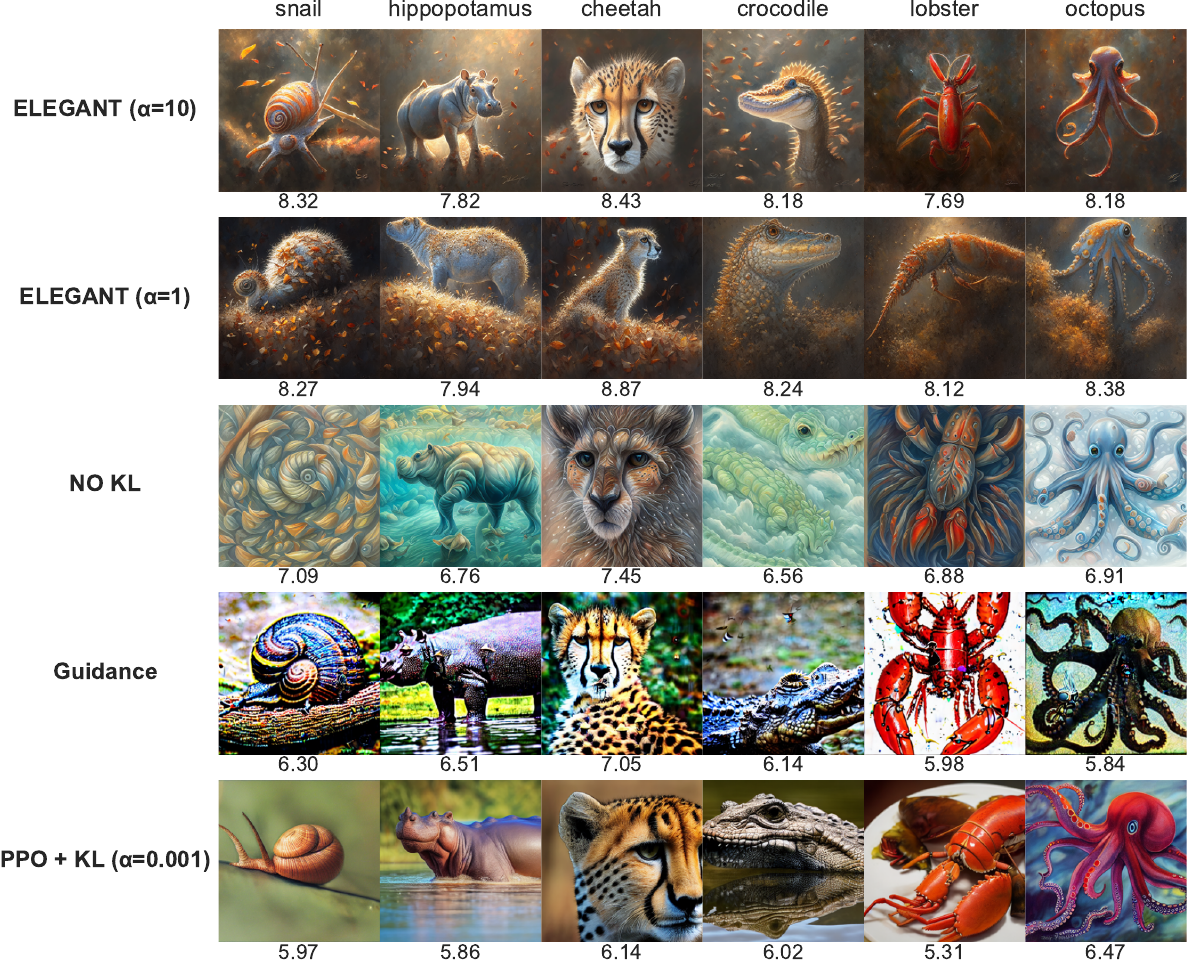}
    \caption{More images generated by \agl and baselines. \textbf{Guidance} is trained on AVA dataset~\citep{murray2012ava}. All other algorithms (\textbf{NO KL}, \textbf{Guidance}, \textbf{PPO + KL}, and \textbf{ELEGANT}) make $12800$ reward inquiries to perform fine-tuning. All evaluation prompts are not seen in training.}
    \label{fig:more-samples}
\end{figure}

In Table~\ref{tab:image}, we report the performances of our \agl and baselines in fine-tuning aesthetic scores.
We provide more qualitative samples to illustrate the performances in Figure~\ref{fig:more-samples}. 
From the visual evidence, \agl with $\alpha=10$ consistently produces images with higher scores compared to the baseline, suggesting an improved capacity for generating high-fidelity images. This improvement is maintained, albeit to a slightly lesser degree, even when the $\alpha$ parameter is reduced to 1.

\paragraph{Compared with \textbf{NO KL}.} Qualitatively, the images generated by \agl are depicted with a high level of detail and realism, which may be indicative of the algorithm's capability to preserve the distinctive features of each species. The \textbf{NO KL} baseline falls short in achieving scores comparable to those of \agl, particularly notable in the representations of the hippopotamus and the crocodile. Moreover, experimentally we observe severe reward collapse as early as the 10th epoch of \textbf{NO KL} (at the time, the reward is $\approx 7$), which indicates that even if the optimization continues (i.e., reward score rises), it is meaningless because \textbf{NO KL} fails to produce meaningful generations afterward. Reward collapse is regarded as a major issue in reward-guided diffusion model fine-tuning~\citep{clark2023directly,prabhudesai2023aligning}.

\paragraph{Compared with \textbf{Guidance}.} In practice, we observe that the guidance strength in \textbf{Guidance} is hard to control: if the guidance level and target level are not strong, the reward-guided generation would be weak (cf. Table~\ref{tab:image_conditional}). However, with a strong guidance signal and a high target value, the generated images
become more colorful at the expense of degradation of the image qualities.  In presenting qualitative results in Figure~\ref{fig:more-samples}, we set the target as $10$ and the guidance level as $200$ to balance generation performance and fidelity.

\begin{figure}[!t]
    \centering
\includegraphics[width=\linewidth]{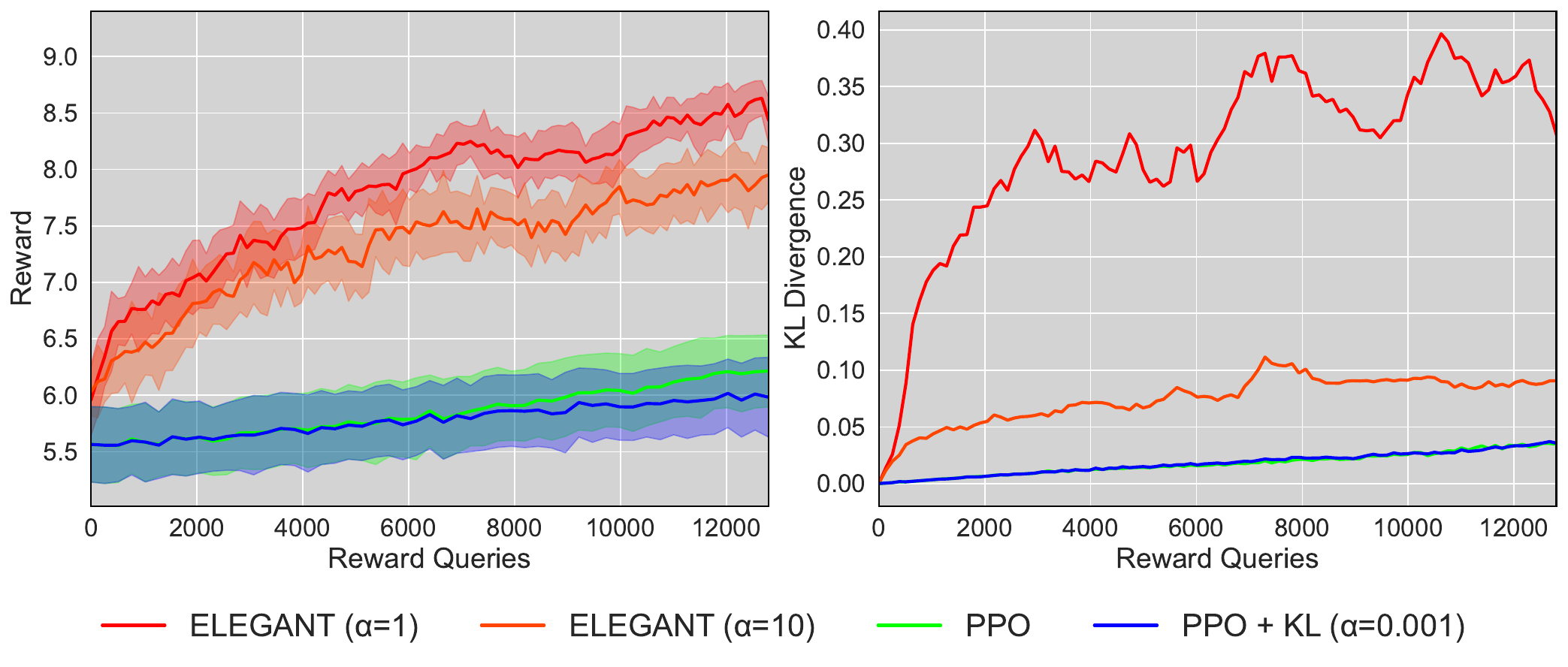}
    \caption{Training curves of reward (left) and KL divergence (right) for \textbf{PPO}, \textbf{PPO + KL}, and \agl for fine-tuning aesthetic scores. The $x$ axis corresponds to the number of reward queries in the fine-tuning process.  } 
    \label{fig:training_curve_ddpo}
\end{figure}

\paragraph{Compared with \textbf{PPO}.}
We plot the training curves of \textbf{PPO}, \textbf{PPO + KL}, versus \agl in Figure~\ref{fig:training_curve_ddpo}.
Empirically, we observe  \textbf{PPO} outperforms \textbf{PPO + KL} in terms of reward but still falls short compared to our \textbf{ELEGANT}.

\paragraph{Compared with \textbf{PPO + KL}.} As a KL-regularized PPO-based algorithm, \textbf{PPO + KL} maintains a stable generation quality and is robust in KL divergence, indicating a stable optimization process. However, with limited training samples, this baseline becomes inefficient in both training time and synthesis performances.

\begin{table}[!t]
    \centering
\caption{Results of classifier guidance for aesthetic scores. ($\cdot$) are confidence intervals. Note the top $5 \%$ value is 6.00. 
}
    \begin{tabular}{p{0.14\linewidth}p{0.16\linewidth} p{0.14\linewidth} p{0.10\linewidth} } \toprule 
Target ($y_{\condition}$) & Guidance level ($\gamma$) &  Reward $(r)$ & KL-Div\\ \midrule    
          7 &100 &  5.69 (0.21)& 0.34\\ 
   10 &100 & 6.02 (0.21)  & 2.45\\
   16& 100 & 6.32 (0.39)& 9.33\\
   10 & 200 & 6.41 (0.19)& 5.99\\   
      
   \bottomrule 
    \end{tabular} 
    \label{tab:image_conditional}
\end{table}

\end{document}